%% file: elsarticle-template-num.tex
\documentclass[preprint,12pt]{elsarticle}
\usepackage{amsfonts}  
\usepackage{amsmath}
\usepackage{xcolor}
\usepackage{amsthm}
\usepackage{multirow}
\usepackage{booktabs}
\usepackage{graphicx}
\usepackage{subcaption}
\usepackage{enumitem}
\usepackage{float}
\usepackage{url} 
\usepackage{longtable}
\usepackage{multirow}
\usepackage{caption}
\usepackage{hyperref}
\usepackage{makecell}
\usepackage{bm}
\usepackage{booktabs}

\begin{document}

\begin{frontmatter}

\title{An Adaptive Physics-Guided Framework for Lithium-Ion Battery Temperature Forecasting}

\author[1]{Khoa~Tran\fnref{equal}\corref{cor1}}
\ead{trandinhkhoa@tdtu.edu.vn}

\author[2]{Phone~Wai~Yan~Moe\fnref{equal}}
\ead{wyan40653@gmail.com}

\author[2]{Tri~Le}
\ead{tri.le@aiware.website}

\cortext[cor1]{Corresponding author:
\href{mailto:trandinhkhoa@tdtu.edu.vn}{trandinhkhoa@tdtu.edu.vn}}

\fntext[equal]{Khoa Tran and Phone Wai Yan Moe contributed equally to this work.}

\affiliation[1]{
    organization={Data Science Laboratory, Faculty of Information Technology, Ton Duc Thang University},
    city={Ho Chi Minh City},
    postcode={70000},
    country={Vietnam}
}

\affiliation[2]{
    organization={AIWARE Limited Company},
    addressline={17 Huynh Man Dat Street, Hoa Cuong Ward, Hai Chau District},
    city={Da Nang},
    postcode={550000},
    country={Vietnam}
}

\begin{abstract}
Accurate forecasting of lithium-ion battery temperature is essential 
for proactive thermal management and safe battery operation. However, 
conventional data-driven approaches relying solely on measured signals 
within finite observation windows may inadequately capture historical 
operating information, inter-variable dependencies, and variations 
across battery cells and operating conditions. To address these 
limitations, this study proposes an adaptive physics-guided framework 
for future battery surface temperature prediction. A second-order 
equivalent circuit model (ECM) incorporating hysteresis is combined 
with an extended Kalman filter (EKF) to estimate physically 
interpretable electrical states, including the state of charge, 
polarisation voltages, and hysteresis state. These states complement 
measured current, voltage, and temperature by providing historical 
electrical context. An inverted Transformer (iTransformer) is 
integrated with a gated recurrent unit (GRU) to jointly capture 
inter-variable dependencies and temporal dynamics. Furthermore, 
low-rank adaptation (LoRA) enables parameter-efficient online 
adaptation to cell-specific characteristics using only previously 
observed temperature targets. The proposed framework is evaluated 
on two experimental datasets comprising LCO and NMC batteries 
under diverse loading profiles and ambient temperatures ranging 
from $5$ to $45\,^{\circ}\mathrm{C}$. With online adaptation, the 
proposed framework achieves root mean squared errors (RMSEs) of 
0.0491, 0.1703, and 0.5867~$^{\circ}$C on the LCO dataset and 
0.0065, 0.1037, and 0.2556~$^{\circ}$C on the NMC dataset for 
prediction horizons of 10, 100, and 300~s, respectively. The 
proposed adaptive framework significantly outperforms all evaluated 
learned baselines at all three horizons on the LCO dataset and 
at the 10~s horizon on the NMC dataset. These findings demonstrate 
the potential of integrating physics-guided state representation, 
channel--temporal learning, and online adaptation for accurate 
and adaptive battery temperature forecasting.

\end{abstract}

\end{frontmatter}

\section{Introduction}

The global energy sector is undergoing a significant transition from fossil fuel-based systems toward cleaner and more sustainable energy technologies to mitigate climate change and reduce greenhouse gas emissions. This transition has accelerated the adoption of electric vehicles (EVs)~\cite{mustafa2026role} and the integration of renewable energy sources, particularly photovoltaic (PV) systems~\cite{dransfield2026residential}. Lithium-ion batteries (LIBs) play a critical role in these applications due to their high energy density, long cycle life, and high energy efficiency~\cite{he2026optimization}. However, their widespread deployment also raises significant safety and reliability concerns, particularly under demanding operating conditions. One of the primary safety challenges associated with LIBs is thermal instability. Excessive heat generation, inadequate heat dissipation, and abnormal operating conditions can accelerate battery degradation and, in severe cases, trigger thermal runaway~\cite{sun2026thermal}, potentially resulting in fires or explosions.

To address these challenges, physics-based, data-driven, and hybrid models have been developed for battery temperature estimation and prediction. Ren et al.~\cite{ren2025temperature} proposed Ada-GRU to improve generalization across operating conditions. Temporal Distribution Characterization (TDC) partitions training sequences by maximizing distributional differences using maximum mean discrepancy (MMD). A GRU then extracts hidden-state representations, while Temporal Distribution Matching (TDM) minimizes their distributional differences through an adaptively weighted loss. This promotes transferable representations across current profiles, ambient temperatures, and battery types. Shen et al.~\cite{shen2025physics} incorporated residuals from a one-dimensional, three-node heat balance model into the loss function of an long short-term memory (LSTM) network to encourage physically consistent blade battery temperature estimates. Using the same limited training dataset, this hybrid model achieved lower RMSE and MAE than a conventional LSTM at all three charging rates tested at 35~$^\circ$C. To address seasonal variations in battery temperature, Ding et al.~\cite{ding2026temperature} proposed a K-means clustering approach based on the daily mean squared error (MSE) between battery and ambient temperatures. The resulting clusters were consolidated into three seasonal partitions (spring/autumn, summer, and winter), with a separate prediction model trained for each. The proposed PINN-GRU--Transformer architecture combines GRU to capture short-term temperature fluctuations and Transformer to model long-term dependencies. Furthermore, a physics-informed loss function derived from the lumped capacitance thermal model is incorporated to enforce physical consistency. Experimental results demonstrate that the proposed approach outperforms purely data-driven models in battery temperature prediction. Li et al.~\cite{li2025multimodal} leverages 2D thermal images and 1D temperature data to predict future battery temperatures during thermal runaway. The two modalities are processed through separate feature extraction branches, consisting of ResNet18–GRU and 1D CNN–GRU, respectively. The extracted features are then aligned using a bidirectional cross-attention mechanism and integrated through a multi-scale gated fusion module to generate temperature predictions. Compared with conventional CNN–GRU, the proposed method reduces MSE and MAE by 91.07\% and 75.45\%, respectively. However, its validation relies on simulation-generated data and a random sample-level dataset split, leaving its generalizability to independent real-world thermal runaway events insufficiently established.

The aforementioned studies rely on measured signals as inputs. To enrich the input representation and improve predictive performance, other studies combine physics-based state estimates with measured features for battery temperature estimation. Yang et al.~\cite{yang2026physics} coupled an ECM with a lumped thermal network and used an extended Kalman filter (EKF)~\cite{ribeiro2004kalman} to estimate SOC and temperature. These estimates were then incorporated into a GRU network as physical priors. However, although the GRU captures temporal dependencies, explicitly modeling interactions among input features may offer further benefits. Moreover, the framework was validated on a pouch cell, leaving its applicability to cylindrical cells unverified. The study also focused on current-state estimation; extending this approach to future temperature prediction could support earlier intervention and proactive thermal management. Teng et al.~\cite{teng2025physics} proposed a physics-guided LSTM for online estimation of lithium-ion battery core temperature. Their multilayer electrothermal coupling model (METCM) used an EKF to update the convective heat-transfer coefficient online based on surface-temperature measurements, allowing the model to account for changing cooling conditions. The LSTM was trained to estimate core-temperature change from five inputs provided by the METCM: core heat generation, heat conduction to each of the two adjacent layers, heat dissipation through the tabs, and state of charge (SOC). On the Dynamic Stress Test profile, the hybrid method reduced RMSE by 32.9\% compared with the adaptive METCM alone.

Although combining physics-based state estimates with measured 
signals has demonstrated improvements in temperature prediction 
accuracy, existing frameworks primarily focus on modelling temporal 
dependencies, with limited attention to interactions among input 
variables. Additionally, battery cell characteristics may vary due 
to manufacturing variability and differences in operating conditions, 
such as ambient temperature. These variations motivate the need 
for online adaptation of the prediction model to maintain predictive 
accuracy during operation.

To address these limitations, this study proposes an adaptive 
physics-guided framework for lithium-ion battery temperature 
forecasting. The main contributions of this study are summarised 
as follows:
\begin{enumerate}

    \item Rather than relying solely on measured current, voltage and 
    temperature ($I$, $V$, $T$), the proposed framework incorporates 
    physically interpretable electrical states, including the state 
    of charge (SOC), polarisation voltages ($U_1$, $U_2$) and hysteresis 
    state ($H$), estimated using a second-order equivalent circuit 
    model (ECM) and an extended Kalman filter (EKF). These states encode 
    historical operating information beyond the finite measurement 
    window, enriching the input representation and improving 
    temperature prediction accuracy.

    \item A hybrid neural architecture integrating an inverted Transformer 
    (iTransformer) and a gated recurrent unit (GRU) is developed to 
    jointly capture inter-variable and temporal dependencies. The 
    iTransformer models interactions among measured variables, and its 
    output is concatenated with the electrical states as contextual 
    information before being processed by the GRU to capture temporal 
    dependencies across consecutive forecasting steps. Ablation results 
    demonstrate the benefits of combining these complementary mechanisms.

    \item A low-rank adaptation (LoRA) mechanism is incorporated to enable 
    continual adaptation to cell-specific characteristics and varying 
    operating conditions without updating the entire neural network. 
    The adaptation follows a causal learning strategy in which only 
    previously observed temperature targets are used for parameter 
    updates, preventing future information leakage. This mechanism 
    improves forecasting accuracy relative to the non-adaptive model 
    while keeping the original network parameters frozen.

    \item The proposed framework is evaluated on two experimental datasets 
    comprising different lithium-ion battery chemistries (LCO and NMC), 
    ambient temperatures ranging from $5$ to $45\,^{\circ}\mathrm{C}$, 
    and diverse loading profiles. Baseline comparisons, 
    ablation studies, statistical significance tests and cross-dataset 
    experiments are conducted to assess the predictive performance, 
    adaptability and generalisability of the proposed architecture.

\end{enumerate}

The remainder of this paper is organised as follows. Section 2 
introduces the proposed methodology. Section 3 describes the 
experimental setup. Section 4 presents and discusses the results. 
Section 5 concludes the study, and Section 6 provides additional 
experimental results.

\input{method}
\input{experiment}

\section{Conclusion}

This study proposed an adaptive physics-guided framework for 
lithium-ion battery surface temperature forecasting. The framework 
integrates ECM-derived electrical states to enrich historical 
operating information, an iTransformer and GRU to jointly capture 
inter-variable dependencies and temporal dynamics, and LoRA to 
enable parameter-efficient online adaptation to varying operating 
conditions. Experimental results on LCO and NMC batteries across 
ambient temperatures of $5$--$45\,^{\circ}\mathrm{C}$ demonstrate 
the benefits of the proposed framework. With online adaptation, 
the proposed method achieves the lowest RMSE among the evaluated 
methods at prediction horizons of 10, 100, and 300~s on the LCO 
dataset and 10 and 100~s on the NMC dataset. These findings highlight 
the potential of integrating physics-guided state representation, 
channel--temporal learning, and continual adaptation for proactive 
battery thermal management. Future work will focus on improving 
computational efficiency, cross-chemistry transferability, and 
extending the framework to pack-level temperature forecasting 
and practical BMS deployment.

\section*{Contact Information}
For access to the code and further information about this proposed system, please contact AIWARE Limited Company at: \url{https://aiware.website}

\input{appendix}

\bibliographystyle{plain}
\bibliography{cas-refs}
\end{document}

%% file: method.tex
\section{Proposed method}
\label{sec:proposed_method}

\subsection{Problem formulation and framework overview}
\label{sec:method_overview}

The objective is to forecast the surface temperature of a lithium-ion cell from sequential measurements of current, terminal voltage and surface temperature. At discrete time step $t$, let $I_t$, $V_t$ and $T_t$ denote these measurements, and let $\Delta t$ denote the sampling interval. Given a lookback of $L$ steps, the measurement history is
\begin{equation}
 \mathbf{X}_t =
 \begin{bmatrix}
 I_{t-L+1} & \cdots & I_t \\
 V_{t-L+1} & \cdots & V_t \\
 T_{t-L+1} & \cdots & T_t
 \end{bmatrix}\in\mathbb{R}^{3\times L}.
 \label{eq:method_input}
\end{equation}
For a horizon of $h$ steps, the task is to estimate $T_{t+h}$, i.e.\ the temperature $h\Delta t$ seconds ahead. No future current, voltage or temperature is supplied to the predictor. Instead of estimating $T_{t+h}$ directly, the method predicts the signed temperature change $\Delta T_{t,h}=T_{t+h}-T_t$ and adds it to the current measured temperature. The forecast is therefore a learned correction to the persistence forecast, obtained in one step without recursive predictions at intermediate steps.

\begin{figure}[t]
\centering
\includegraphics[width=\textwidth]{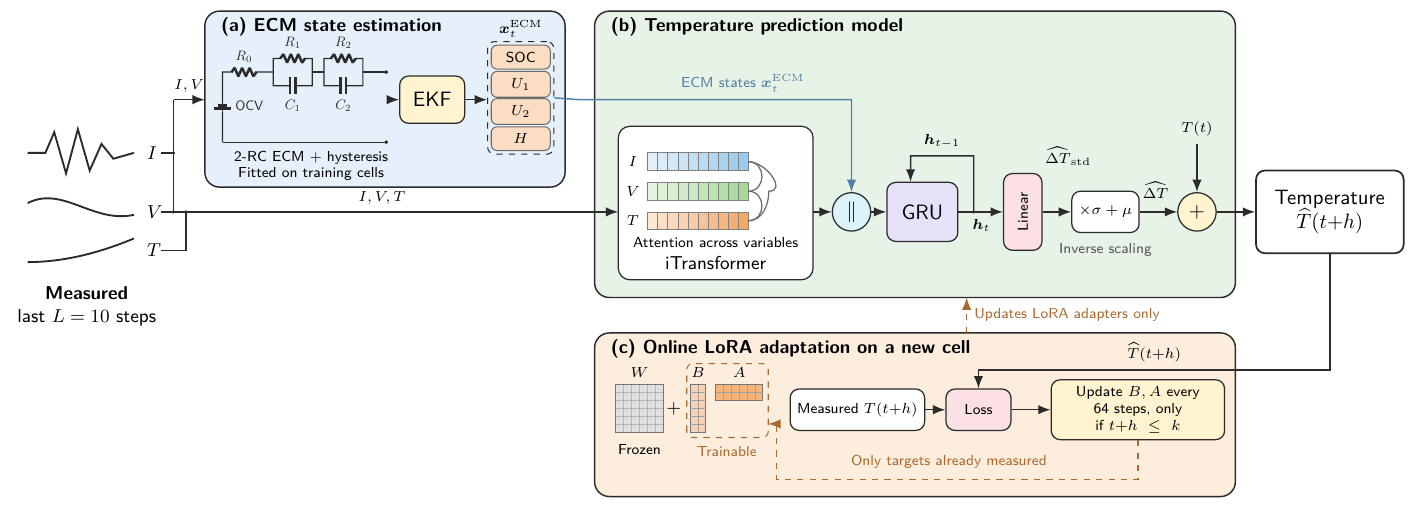}
\caption{Overview of the proposed framework. (a) A 2-RC ECM with hysteresis and an EKF estimate the electrical state $\mathbf{c}_t$. (b) An iTransformer encodes the last $L$ samples of $I$, $V$ and $T$; the result is fused with $\mathbf{c}_t$ in a GRU cell, which predicts the temperature change added to $T_t$. (c) On a new cell, LoRA adapters are updated online using only measured targets ($t+h\le k$).}
\label{fig:architecture}
\end{figure}

The framework, shown in Fig.~\ref{fig:architecture}, has three components. (a) A second-order equivalent circuit model (2-RC ECM) with an exponential open-circuit voltage (OCV) function and one-state hysteresis is combined with an extended Kalman filter (EKF) to estimate the electrical state from current and voltage. (b) An inverted Transformer (iTransformer)~\cite{liu2024itransformer} encodes the measured history with self-attention across measurement channels; its output is concatenated with the current ECM state and passed to a gated recurrent unit (GRU) cell~\cite{cho2014learning}, whose hidden state is carried between consecutive forecasting steps, and a linear layer outputs the temperature change. (c) Low-rank adaptation (LoRA)~\cite{hu2022lora} updates the trained predictor on a new cell during operation, using only temperatures that have already been measured, while the original weights stay frozen.

The ECM supplies physically interpretable information about the internal electrical state, and the neural predictor learns how this state and the measured signals relate to the subsequent temperature change. The final framework does not impose an explicit thermal balance equation, does not use the ambient temperature as an input and does not add a physics-residual term to the training loss; physics enters the predictor only through the ECM states. The predictor is evaluated in two forms: the trained model without adaptation (Proposed) and the same model with online adaptation (Proposed + LoRA). The final configuration is listed in Table~\ref{tab:hyperparameters}.

\subsection{Electrical state estimation}
\label{sec:method_ecm}

\subsubsection{Second-order equivalent circuit model}

The ECM consists of an OCV source, an ohmic resistance $R_0$ and two parallel resistor--capacitor branches $(R_1,C_1)$ and $(R_2,C_2)$, augmented with a hysteresis state. The state vector is
\begin{equation}
 \mathbf{x}_t = [s_t, U_{1,t}, U_{2,t}, H_t]^{\mathsf{T}},
 \label{eq:method_ecm_state}
\end{equation}
where $s_t\in[0,1]$ is the state of charge (SOC), $U_{1,t}$ and $U_{2,t}$ are the polarisation voltages and $H_t\in[-1,1]$ is the normalised hysteresis state. Discharge current is positive. Because the current measured at step $t-1$ acts over the interval from $t-1$ to $t$, the model uses $i_t=I_{t-1}$. The SOC follows coulomb counting, and the RC voltages are discretised with the bilinear (Tustin) transform using the average current over the interval:
\begin{align}
 s_t &= \operatorname{clip}_{[0,1]}\!\left(s_{t-1}-\frac{\eta\,\Delta t}{Q_n}\,i_t\right),
 \label{eq:method_soc}\\
 U_{j,t} &= a_j U_{j,t-1}+b_j\,\frac{i_t+i_{t-1}}{2},
 \qquad j\in\{1,2\},
 \label{eq:method_rc}\\
 a_j &= \frac{1-\Delta t/(2\tau_j)}{1+\Delta t/(2\tau_j)},\qquad
 b_j = \frac{\Delta t/C_j}{1+\Delta t/(2\tau_j)},\qquad \tau_j=R_jC_j,
 \label{eq:method_rc_coeff}
\end{align}
where $Q_n=3600\,Q_{\mathrm{cell}}$ is the nominal capacity in ampere-seconds ($Q_{\mathrm{cell}}=2.5$ and 4.93\,Ah for the first and second datasets), $\Delta t$ is in seconds and the coulombic efficiency is $\eta=1$.

The OCV is the exponential--polynomial function of Chen and Rinc\'on-Mora~\cite{chen2006accurate},
\begin{equation}
 V_{\mathrm{OC}}(s)=\beta_0 e^{-\beta_1 s}+\beta_2+\beta_3 s+\beta_4 s^2+\beta_5 s^3,
 \label{eq:method_ocv}
\end{equation}
with coefficients $\bm{\beta}=(\beta_0,\dots,\beta_5)$. Hysteresis follows the one-state model of Plett~\cite{plett2004ekf2}: the state decays towards $-1$ during discharge and towards $+1$ during charge at a rate proportional to the charge throughput,
\begin{equation}
 H_t=\kappa_t H_{t-1}-(1-\kappa_t)\operatorname{sgn}(i_t),
 \qquad
 \kappa_t=\exp\!\left(-\frac{\eta\,\gamma_H\,|i_t|\,\Delta t}{Q_n}\right),
 \label{eq:method_hyst}
\end{equation}
where $\gamma_H$ is the hysteresis rate. The predicted terminal voltage is
\begin{equation}
 \widehat V_t=g(\mathbf{x}_t,i_t)=V_{\mathrm{OC}}(s_t)-R_0 i_t-U_{1,t}-U_{2,t}+M_H H_t,
 \label{eq:method_voltage}
\end{equation}
where $M_H$ is the maximum hysteresis voltage. The selected ECM is electrical only: its resistances are not coupled to the measured temperature. The OCV function, RC order, hysteresis and temperature-dependent (Arrhenius) resistances are compared in Appendix~\ref{app:ecm}.

\subsubsection{ECM parameter identification}

The parameters $\bm{\vartheta}=\{R_0,R_1,R_2,C_1,C_2,M_H,\gamma_H,\bm{\beta}\}$ are identified separately for each dataset on its training split, by simulating Eqs.~\eqref{eq:method_soc}--\eqref{eq:method_voltage} in open loop (without the filter) from the reference initial SOC of each sequence and minimising
\begin{equation}
 \mathcal{L}_{\mathrm{ECM}}(\bm{\vartheta})=
 \frac{1}{N_{\mathrm{tr}}}
 \sum_{t\in\mathcal{D}_{\mathrm{tr}}}
 \left[(V_t-\widehat V_t)^2
 +\lambda_s(s_t^{\mathrm{ref}}-\widehat s_t)^2\right],
 \qquad \lambda_s=0.1,
 \label{eq:method_ecm_loss}
\end{equation}
where $\mathcal{D}_{\mathrm{tr}}$ contains all simulated steps of the training sequences, $N_{\mathrm{tr}}=|\mathcal{D}_{\mathrm{tr}}|$, voltage is in volts and SOC is a fraction. The reference SOC $s_t^{\mathrm{ref}}$ is obtained by coulomb counting. The objective is minimised with L-BFGS-B within box bounds. All parameters are mapped to $[0,1]$ for optimisation, with logarithmic scaling for $R_j$, $C_j$, $M_H$ and $\gamma_H$, so that resistances of order $10^{-2}$\,$\Omega$ and capacitances of order $10^{4}$\,F receive comparable steps. The OCV coefficients are initialised by least squares on the training SOC--voltage pairs and are then fitted jointly with the circuit parameters. The ECM is not refitted on validation or test data. The optimiser settings are given in Section~\ref{sec:experiment_implementation}.

\subsubsection{Extended Kalman filter}

During temperature prediction, the ECM states are estimated with an EKF, which corrects the open-loop states with the measured voltage. For a given current, Eqs.~\eqref{eq:method_soc}--\eqref{eq:method_hyst} are linear in the state (apart from clipping), so the state-space model is
\begin{align}
 \mathbf{x}_t &= \mathbf{F}_t\mathbf{x}_{t-1}+\mathbf{d}_t+\mathbf{w}_t,
 \qquad \mathbf{F}_t=\operatorname{diag}(1,a_1,a_2,\kappa_t),\\
 V_t &= g(\mathbf{x}_t,i_t)+v_t,
 \label{eq:method_state_space}
\end{align}
where $\mathbf{d}_t$ collects the current-dependent terms of Eqs.~\eqref{eq:method_soc}--\eqref{eq:method_hyst}, and $\mathbf{w}_t$ and $v_t$ are process and measurement noise with covariances $\mathbf{Q}_w$ and $R_V$. The prediction step is
\begin{align}
 \widehat{\mathbf{x}}_{t|t-1} &= \mathbf{F}_t\widehat{\mathbf{x}}_{t-1|t-1}+\mathbf{d}_t,\\
 \mathbf{P}_{t|t-1} &= \mathbf{F}_t\mathbf{P}_{t-1|t-1}\mathbf{F}_t^{\mathsf{T}}+\mathbf{Q}_w.
 \label{eq:method_ekf_predict}
\end{align}
The nonlinearity is in the OCV, so the observation Jacobian is
\begin{equation}
 \mathbf{J}_t=\left[\frac{\mathrm{d}V_{\mathrm{OC}}}{\mathrm{d}s}\Big|_{\widehat s_{t|t-1}},\,-1,\,-1,\,M_H\right],
 \label{eq:method_jacobian}
\end{equation}
with the OCV slope evaluated by a central difference ($\delta s=10^{-4}$). The innovation, Kalman gain and posterior update are
\begin{align}
 e_t &= V_t-g(\widehat{\mathbf{x}}_{t|t-1},i_t),\\
 \mathbf{K}_t &= \mathbf{P}_{t|t-1}\mathbf{J}_t^{\mathsf{T}}
 \left(\mathbf{J}_t\mathbf{P}_{t|t-1}\mathbf{J}_t^{\mathsf{T}}+R_V\right)^{-1},\\
 \widehat{\mathbf{x}}_{t|t} &= \widehat{\mathbf{x}}_{t|t-1}+\mathbf{K}_t e_t,\qquad
 \mathbf{P}_{t|t} = (\mathbf{I}_4-\mathbf{K}_t\mathbf{J}_t)\mathbf{P}_{t|t-1},
 \label{eq:method_ekf_update}
\end{align}
after which SOC is clipped to $[0,1]$ and $H$ to $[-1,1]$. The noise covariances are $\mathbf{Q}_w=\operatorname{diag}(10^{-10},10^{-6},10^{-6},10^{-6})$ and $R_V=0.15^2$\,V$^2$, so coulomb counting is trusted, and the initial covariance is $\mathbf{P}_0=10^{-4}\mathbf{I}_4$. Each sequence starts from $\widehat{\mathbf{x}}_0=[s_0^{\mathrm{ref}},0,0,H_0]^{\mathsf{T}}$, with $H_0=-1$ for the first dataset, whose cycles start after a discharge, and $H_0=+1$ for the second, whose profiles start after a full charge. The reported ECM voltage error uses the prior prediction $g(\widehat{\mathbf{x}}_{t|t-1},i_t)$, made before $V_t$ is assimilated. The electrical context passed to the temperature predictor is the posterior state at the forecast origin,
\begin{equation}
 \mathbf{c}_t=\widehat{\mathbf{x}}_{t|t}\in\mathbb{R}^{4}.
 \label{eq:method_context}
\end{equation}
This context describes the electrical state at the forecast origin rather than a window of estimated states. The KF, EKF, UKF and particle filter are compared in Table~\ref{tab:filter_ablation}, and the use of the current state instead of a window of ECM states in Table~\ref{tab:feature_ablation}.

\subsection{Recurrent iTransformer with ECM context}
\label{sec:method_predictor}

\subsubsection{Inverted channel tokenisation and attention}
\label{sec:method_channel}

The measurement window and the ECM context carry complementary information: the window describes the recent evolution of the three measured signals, and the context summarises the current electrical state. The predictor first encodes the window and then fuses the result with the context.

Let $\widetilde{\mathbf{X}}_t$ and $\widetilde{\mathbf{c}}_t$ denote the window and the context standardised with the mean and standard deviation of each input feature on the training data. Unlike temporal tokenisation, the iTransformer~\cite{liu2024itransformer} treats the whole lookback series of each variable as one token. A linear embedding shared by the three channels gives
\begin{equation}
 \mathbf{E}_t=\widetilde{\mathbf{X}}_t\mathbf{W}_e+\mathbf{1}\mathbf{b}_e^{\mathsf{T}}
 \in\mathbb{R}^{3\times d},
 \label{eq:method_embedding}
\end{equation}
where $\mathbf{W}_e\in\mathbb{R}^{L\times d}$ and $\mathbf{b}_e\in\mathbb{R}^{d}$, with $L=10$ and $d=32$. Self-attention therefore operates on three tokens, one per measured variable. For head $m$,
\begin{equation}
 \mathbf{A}_t^{(m)}=\operatorname{softmax}\left(
 \frac{\mathbf{E}_t\mathbf{W}_q^{(m)}\bigl(\mathbf{E}_t\mathbf{W}_k^{(m)}\bigr)^{\mathsf{T}}}{\sqrt{d_a}}
 \right)\mathbf{E}_t\mathbf{W}_v^{(m)},
 \qquad d_a=d/M,
 \label{eq:method_attention}
\end{equation}
where $M$ is the number of heads and the softmax is taken over the key tokens. The encoder has one layer with $M=2$ heads ($d_a=16$), so each head forms a $3\times3$ attention map, and the representation of each signal depends on the other two within the same window. The head outputs are concatenated and projected back to dimension $d$. The layer follows the standard post-normalisation Transformer encoder, with residual connections and layer normalisation after the attention and after a two-layer feed-forward network of width $2d=64$ with ReLU activation, and dropout 0.1. No positional or channel embedding is added; the channels keep their identity through the fixed flattening order. The encoder output is
\begin{equation}
 \mathbf{Z}_t=\operatorname{Encoder}(\mathbf{E}_t)\in\mathbb{R}^{3\times32},
 \qquad \mathbf{z}_t=\operatorname{vec}(\mathbf{Z}_t)\in\mathbb{R}^{96},
 \label{eq:method_encoder}
\end{equation}
where $\operatorname{vec}$ concatenates the $I$, $V$ and $T$ tokens in this order. The lookback length is studied in Appendix~\ref{app:lookback}, and temporal, channel and combined encoders are compared in Table~\ref{tab:temporal_channel_ablation}.

\subsubsection{ECM context fusion and recurrent hidden state}
\label{sec:method_recurrence}

A finite window limits the history available at each step. To retain information across successive windows, a GRU cell follows the channel encoder. The encoded window is concatenated with the standardised ECM context,
\begin{equation}
 \mathbf{u}_t=[\mathbf{z}_t;\widetilde{\mathbf{c}}_t]\in\mathbb{R}^{100},
 \label{eq:method_fusion}
\end{equation}
so the four ECM states enter as context rather than as additional attention tokens. Let $\mathbf{m}_t\in\mathbb{R}^{64}$ denote the recurrent hidden state (written $\mathbf{m}$ to avoid confusion with the horizon $h$ and the hysteresis state $H$). It is updated as
\begin{equation}
 \mathbf{m}_t=\operatorname{GRUCell}(\mathbf{u}_t,\mathbf{m}_{t-1}),
 \label{eq:method_gru}
\end{equation}
where the update and reset gates of the GRU~\cite{cho2014learning} control how much of the new input is written into the state and how much of the previous state is kept. The hidden state is set to zero at the start of each cycle or profile, the samples are processed in chronological order, and $\mathbf{m}_t$ is passed to the next forecasting step. This state is distinct from the EKF state and has no direct physical meaning. Appendix~\ref{app:hidden} compares the gated update with a non-gated recurrent update and with the non-recurrent iTransformer.

\subsubsection{Temperature-change prediction}
\label{sec:method_target}

A linear layer maps the hidden state to the standardised temperature change,
\begin{equation}
 \widehat y_{t,h}=\mathbf{w}_o^{\mathsf{T}}\mathbf{m}_t+b_o,
 \label{eq:method_output}
\end{equation}
with $\mathbf{w}_o\in\mathbb{R}^{64}$ and $b_o\in\mathbb{R}$. The training target is
\begin{equation}
 y_{t,h}=\frac{T_{t+h}-T_t-\mu_h}{\sigma_h},
 \label{eq:method_target}
\end{equation}
where $\mu_h$ and $\sigma_h$ are the mean and standard deviation of $T_{t+h}-T_t$ over the training samples at horizon $h$. The forecast in degrees Celsius is
\begin{equation}
 \widehat T_{t+h}=T_t+\mu_h+\sigma_h\widehat y_{t,h}.
 \label{eq:method_forecast}
\end{equation}
The target is the signed temperature difference, not the relative change $(T_{t+h}-T_t)/T_t$ or the temperature itself. No thermal-model residual or heat-generation feature is used in the final predictor. These alternatives are compared in Appendix~\ref{app:target} (Tables~\ref{tab:physics_ablation} and~\ref{tab:target_comparison}).

\subsection{Offline training and causal online adaptation}
\label{sec:method_training}

\subsubsection{Offline training}
\label{sec:method_offline}

The ECM is identified first, and the EKF states of all sequences are then used as inputs to the temperature predictor. Within each cycle or profile of $N$ samples, a forecast is made at every origin $t$ with a complete window and a target inside the same sequence, i.e.\ $L-1\le t\le N-1-h$ (samples indexed from 0); windows never cross sequence boundaries. For horizon $h$, the network parameters $\bm{\theta}$ are trained with the mean squared error of the standardised target,
\begin{equation}
 \mathcal{L}_{\mathrm{temp}}(\bm{\theta})=
 \frac{1}{|\mathcal{S}_h|}\sum_{t\in\mathcal{S}_h}
 (\widehat y_{t,h}-y_{t,h})^2,
 \label{eq:method_train_loss}
\end{equation}
where $\mathcal{S}_h$ is the set of these forecast origins in the training sequences. The hidden state is propagated in time order within each sequence, and training uses truncated backpropagation through time over chunks of 64 steps with 16 cycles per batch. The optimiser and the validation-based stopping are described in Section~\ref{sec:experiment_implementation}. The trained network is the non-adaptive version of the method (Proposed), and it is also the starting point of the adaptive version.

\subsubsection{Low-rank adaptation}
\label{sec:method_lora}

To adapt the trained predictor to a new cell without updating all of its weights, LoRA~\cite{hu2022lora} adds a trainable low-rank correction to selected weight matrices. For a pretrained matrix $\mathbf{W}_0\in\mathbb{R}^{d_{\mathrm{out}}\times d_{\mathrm{in}}}$, the effective weight is
\begin{equation}
 \mathbf{W}=\mathbf{W}_0+\frac{\alpha}{r}\mathbf{B}\mathbf{A},
 \label{eq:method_lora}
\end{equation}
where $\mathbf{A}\in\mathbb{R}^{r\times d_{\mathrm{in}}}$ and $\mathbf{B}\in\mathbb{R}^{d_{\mathrm{out}}\times r}$ are trainable, $r$ is the rank and $\alpha$ is a scaling factor. Adapters are added to the weight matrices of every linear layer (the channel embedding, the attention output projection, both feed-forward layers and the output layer) and to the GRU input-to-hidden and hidden-to-hidden matrices. The query, key and value projections, all biases and the layer-normalisation parameters are not adapted. With $r=8$ and $\alpha=16$, this gives 7,288 trainable parameters, compared with 40,833 in the full network. $\mathbf{A}$ is initialised from $\mathcal{N}(0,0.01^2)$ and $\mathbf{B}$ with zeros, so the adapted predictor is identical to the trained one before the first update. All original weights stay frozen, and the adapters are optimised with AdamW (learning rate $10^{-3}$, weight decay 0.1). The final predictor does not contain a key--value memory layer. LoRA is compared with full fine-tuning and memory-based adaptation in Section~\ref{sec:continual_experiments} (Table~\ref{tab:continual_ablation}).

\subsubsection{Delayed supervision during online inference}
\label{sec:method_online}

A forecast made at time $t$ cannot be checked until $T_{t+h}$ is measured at step $t+h$, so adaptation must separate the time a forecast is issued from the time its target becomes known. Each test cell starts from the trained predictor with new adapters, and its cycles are processed in chronological order in blocks of 64 steps. For each block, the forecasts are first issued with the current model and recorded; these are the reported predictions. One update is then made before the next block, whose first forecast origin is step $k$. The update uses the samples that have not yet been used and whose targets are already measured,
\begin{equation}
 \mathcal{E}_k=\{t\notin\mathcal{U}_k:\ t+h\le k\},
 \label{eq:method_eligible}
\end{equation}
where $\mathcal{U}_k$ is the set of samples used in earlier updates. One AdamW step is taken on
\begin{equation}
 \mathcal{L}_{\mathrm{online},k}(\bm{\phi})=
 \frac{1}{|\mathcal{E}_k|}\sum_{t\in\mathcal{E}_k}
 \bigl(\widehat y_{t,h}(\bm{\phi})-y_{t,h}\bigr)^2,
 \label{eq:method_online_loss}
\end{equation}
where $\bm{\phi}$ contains the LoRA parameters. If $\mathcal{E}_k$ is empty, the update is skipped. For the update, the hidden states of $\mathcal{E}_k$ are recomputed with the current adapters, starting from the hidden state stored during prediction. At the end of each cycle, all remaining samples are used, since their targets lie inside the cycle and have been measured. Each sample is used in exactly one update, so no samples are replayed.

Recorded forecasts are never revised; the measured targets are only used to update the model for later forecasts. The hidden state is reset at each cycle boundary, while the adapters carry over to later cycles of the same cell. The ECM is not updated during adaptation. No update uses a target that had not been measured at the time of the forecast it influences, and no future current, voltage or temperature is used. The horizon study with adaptation in Appendix~\ref{app:horizon_adapt} uses the earlier relative-change target, whereas the main comparison in Table~\ref{tab:main_comparison} uses the target of Eq.~\eqref{eq:method_target}. The same procedure is applied to the second dataset (Section~\ref{sec:second_dataset}) and at the original sampling interval (Section~\ref{sec:original_sampling}).

%% file: experiment.tex
\section{Experimental setup}
\label{sec:experiments}

This section describes the datasets, the evaluation protocol, the implementation, and the baseline methods. The results are presented in Section~\ref{sec:results}.

\subsection{Datasets and data partitioning}
\label{sec:datasets}
\subsubsection{First dataset}
The first dataset comprises laboratory measurements of LCO cells (LCO\_1--6) cycled at ambient temperatures of 5, 15, 25 and 35\,$^\circ$C. The cells have a nominal capacity of 2.5\,Ah. Each cycle consists of constant-current (CC) charging at 2.5\,A to 4.2\,V, constant-voltage (CV) charging at 4.2\,V, CC discharging to 2.0\,V and a rest period. The discharge current is 5\,A in most recordings and 2.5\,A in the remaining ones. The raw records are irregularly spaced, with most intervals close to 10\,s, and are therefore interpolated to a uniform 10\,s grid. LCO\_1--3 are used for training (129 cycles), LCO\_4 for validation (40 cycles), and LCO\_5--6 for testing (82 cycles). This cell-wise partition is used for the model-development experiments.

\subsubsection{Second dataset}
The public drive-cycle dataset of Yao and Kowal~\cite{yao2025multiscale} comprises three LG INR21700 M50LT NMC cells with a capacity of 4.93\,Ah and 12 drive cycles at ambient temperatures of 5, 15, 25, 35 and 45\,$^\circ$C. Each profile consists of a discharge from 100\,\% SOC to 2.5\,V. Measurements acquired at 10\,Hz are averaged over 10\,s for the main experiments; the original sampling interval is examined separately in Section~\ref{sec:original_sampling}. At each ambient temperature, profile IDs 1--5 are assigned to training, IDs 6--7 to validation and IDs 8--12 to testing, resulting in 25 training, 10 validation and 25 test profiles. IDs 9--12 belong to NMC069, which is not used for training, whereas ID 8 belongs to NMC063, a training cell. Consequently, this partition is profile-disjoint but not entirely cell-disjoint. The configuration selected on the first dataset is applied to the second without further configuration tuning. Table~\ref{tab:datasets} summarises both datasets.

\begin{table}[H]
\centering
\caption{Summary of the two datasets.}
\label{tab:datasets}
\resizebox{\textwidth}{!}{%
\begin{tabular}{lll}
\toprule
 & \textbf{First dataset} & \textbf{Second dataset} \\
\midrule
Cell chemistry & LCO & NMC (LG INR21700 M50LT) \\
Nominal capacity & 2.5\,Ah & 4.93\,Ah \\
Voltage window & 2.0--4.2\,V & 2.5--4.2\,V \\
Load profile & CC--CV charge, CC discharge (2.5 or 5\,A) & 12 drive cycles, discharge from 100\,\% SOC \\
Ambient temperatures & 5, 15, 25, 35\,$^\circ$C & 5, 15, 25, 35, 45\,$^\circ$C \\
Measured signals & Current, voltage, temperature & Current, voltage, surface temperature \\
Raw sampling & Irregular, mostly $\approx$10\,s & 10\,Hz \\
Model time step & 10\,s (interpolated) & 10\,s (averaged); 0.1\,s in Section~\ref{sec:original_sampling} \\
Train / validation / test & LCO\_1--3 / LCO\_4 / LCO\_5--6 & Profile IDs 1--5 / 6--7 / 8--12 \\
Number of units & 129 / 40 / 82 cycles & 25 / 10 / 25 profiles \\
\bottomrule
\end{tabular}%
}
\end{table}

\subsection{Evaluation metrics and statistical analysis}
\label{sec:metrics}
Performance is evaluated using root mean squared error (RMSE), mean absolute error (MAE) and the coefficient of determination ($R^2$):
\begin{equation}
\mathrm{RMSE}=\sqrt{\frac{1}{N}\sum_{i=1}^{N}\left(y_i-\hat{y}_i\right)^2},\qquad
\mathrm{MAE}=\frac{1}{N}\sum_{i=1}^{N}\left|y_i-\hat{y}_i\right|,
\label{eq:rmse_mae}
\end{equation}
\begin{equation}
R^2=1-\frac{\sum_{i=1}^{N}\left(y_i-\hat{y}_i\right)^2}{\sum_{i=1}^{N}\left(y_i-\bar{y}\right)^2},
\label{eq:r2}
\end{equation}
where $y_i$ and $\hat{y}_i$ are the measured and predicted values, $\bar{y}$ is the mean of the measured values and $N$ is the number of evaluated samples. Temperature errors are reported in $^\circ$C, voltage errors in mV and SOC errors in percent. Lower RMSE and MAE and higher $R^2$ indicate better predictive performance; temperature RMSE is the primary criterion for configuration comparison. For temperature prediction, RMSE, MAE and $R^2$ are computed over all test samples; $\pm$ denotes the standard deviation over 1000 bootstrap resamples of the test samples. For the ECM, voltage and SOC metrics, including $R^2$, are computed for each test cycle (or profile) and reported as mean $\pm$ standard deviation over test cycles (or profiles). The reference SOC is obtained by coulomb counting.

Persistence is defined by $\widehat{T}_{t+h}=T_t$, where $h$ denotes the prediction horizon. Within a given horizon, methods in the horizon comparisons are evaluated on the same available test samples. For the comparisons employing both statistical criteria, a reduction is considered significant when the 95\,\% bootstrap confidence interval (CI) of the RMSE difference lies below zero and the paired Wilcoxon signed-rank test yields $p<0.05$. The CI is obtained with the percentile method from 2000 bootstrap resamples, and the Wilcoxon test is applied to the per-cycle (or per-profile) RMSE values. The resampling and paired comparison units are test cycles on the first dataset and test profiles on the second. Both criteria must be satisfied; where specified, Holm correction is applied to the Wilcoxon $p$-values.

For the online-adaptation experiments, forgetting is assessed through the change in validation MSE of the standardised training target before and after adaptation, averaged over the two test cells. Because target standardisation depends on the target definition and horizon, these MSE values are compared only within the same target and horizon. The random-seed experiments instead report the mean and sample standard deviation of the RMSE across seeds. No bootstrap CIs or significance tests are available for the original-sampling-interval study.

\subsection{Implementation details}
\label{sec:experiment_implementation}
Unless otherwise stated, the lookback is 10 steps on the 10\,s grid. The initial ablations use a one-step horizon; the main baseline comparisons use $h=1$, 10 and 30 steps (10, 100 and 300\,s). The broader horizon study additionally considers 3, 5, 15 and 60 steps. The early ablations use XGBoost with the relative temperature-change target, $(T_{t+h}-T_t)/T_t$; the final baseline comparisons use the absolute temperature change $\Delta T=T_{t+h}-T_t$. In the target comparison, targets are standardised using training-cell statistics and predictions are converted back to $^\circ$C for evaluation. The measured-input baseline comparisons use training-only scalers. The ECM parameters are fitted on the training cells only by minimising the mean squared voltage error plus 0.1 times the mean squared SOC error over all training samples, using L-BFGS-B (at most 300 iterations) with resistances and capacitances optimised on a logarithmic scale. At test time, the state estimator starts from the reference initial SOC, with initial covariance $10^{-4}$, process-noise variances $10^{-10}$ (SOC), $10^{-6}$\,V$^2$ (RC voltages) and $10^{-6}$ (hysteresis state), and measurement-noise variance $0.15^2$\,V$^2$. XGBoost uses 400 trees, maximum depth 6, learning rate 0.03, and row and column subsampling of 0.9. Neural networks are trained with the mean squared error loss on inputs and targets standardised with training-set statistics.

For the regression-model comparison, neural networks use a hidden size of 64 and two layers, with four attention heads in the Transformer. Training uses Adam with learning rate $10^{-3}$, batch size 1024 and a maximum of 200 epochs, with early stopping on LCO\_4 (patience 20) and seed 42. ElasticNet uses $\alpha=10^{-3}$ and an L1 ratio of 0.5, and the random forest contains 300 trees. The simplified iTransformer uses one layer, two attention heads and embedding dimension 32; TSMixer uses two blocks. The recurrent hidden-state variants are trained with 16 cycles per batch and truncated backpropagation through time over chunks of 64 steps. Training runs for at most 300 epochs, with the learning rate halved after 10 epochs without validation improvement and early stopping with patience 30, no earlier than epoch 100. Other training settings follow the regression-model comparison. The hidden state is reset to zero at the start of each cycle in training, validation and testing, and samples within each cycle are processed in chronological order. Models without the recurrent hidden state follow the original regression-model training protocol. The measured-input neural baselines use a 200-epoch budget and patience 20. Table~\ref{tab:hyperparameters} lists the settings of the final configuration.

\begin{table}[H]
\centering
\caption{Settings of the final configuration.}
\label{tab:hyperparameters}
\begin{tabular}{lp{0.62\textwidth}}
\toprule
\textbf{Setting} & \textbf{Value} \\
\midrule
Lookback / time step & 10 steps / 10\,s \\
Channel tokens & $I$, $V$, $T$ (each embedded from 10 to 32 dimensions) \\
Transformer encoder & 1 layer, 2 heads, $d=32$ \\
ECM context & SOC, $U_1$, $U_2$, $H$ \\
Recurrent cell & GRU cell, input size 100, hidden size 64 \\
Output & Linear layer, standardised $\Delta T$ \\
Trainable parameters & 40\,833 \\
Loss & Mean squared error \\
Optimiser & Adam, learning rate $10^{-3}$, halved after 10 epochs without improvement \\
Batch / BPTT chunk & 16 cycles / 64 steps \\
Epochs & At most 300, minimum 100, early-stopping patience 30 \\
Online adaptation & LoRA ($r=8$, $\alpha=16$), AdamW (learning rate $10^{-3}$, weight decay 0.1), update every 64 steps \\
\bottomrule
\end{tabular}
\end{table}

The ECM, input-feature and lookback ablations were run on one machine, whereas the subsequent horizon and architecture ablations were performed on a different machine equipped with an NVIDIA RTX 5060 GPU. Neural-model inference in the regression comparison is performed on the GPU and the other regression models run on the CPU. Timing comparisons for these ablations are restricted to individual tables. Unless otherwise stated, test times are medians of five runs. ECM time includes state estimation over the test set, and prediction time includes temperature inference and, where applicable, online updates. For the recurrent hidden-state timing study, test cycles are processed in parallel, with sequential processing within each cycle. Measured-input baseline times at the 10\,s interval are single-run timings without ECM estimation; the proposed-model times include ECM estimation and, for LoRA, online updates. Direct-target experiments in Table~\ref{tab:target_comparison} also use single-run timing. Per-step prediction time is the prediction time divided by the number of test samples. All models are implemented in Python using PyTorch, XGBoost, scikit-learn and SciPy.

\subsection{Baseline methods}
\label{sec:baselines}
The final configuration is compared with XGBoost, LSTM, GRU and Transformer using measured $I,V,T$ windows only, without ECM states. The learned baselines use the $\Delta T$ target, the same lookback of 10 and the same dataset partitions. Persistence is included as a non-trained reference. The proposed architecture is evaluated both without and with LoRA online adaptation. Thus, the comparisons assess the combined effect of the architecture, ECM context and, for the adapted variant, online updates, rather than isolating any one of these differences. The proposed method uses the final configuration obtained in the ablation study (Section~\ref{sec:first_dataset}; Table~\ref{tab:hyperparameters}): the electrical 2-RC ECM with exponential OCV, hysteresis and EKF, the iTransformer with GRU hidden state, a 10-step measured-input window and the $\Delta T$ target.

\newcommand{\ecmheader}[1]{%
\multirow{2}{*}{\textbf{#1}} &
\multicolumn{3}{c}{\textbf{Voltage Prediction (mV)}} &
\multicolumn{3}{c}{\textbf{SOC Prediction (\%)}} &
\multicolumn{3}{c}{\textbf{Temperature Prediction ($^\circ$C)}} &
\multirow{2}{*}{\textbf{Time (s)}} \\
\cmidrule(lr){2-4}
\cmidrule(lr){5-7}
\cmidrule(lr){8-10}
&
\textbf{RMSE $\downarrow$} & \textbf{\(R^2\) $\uparrow$} & \textbf{MAE $\downarrow$} &
\textbf{RMSE $\downarrow$} & \textbf{\(R^2\) $\uparrow$} & \textbf{MAE $\downarrow$} &
\textbf{RMSE $\downarrow$} & \textbf{\(R^2\) $\uparrow$} & \textbf{MAE $\downarrow$} &
\\}

\newcommand{\nnheader}[1]{%
\multirow{2}{*}{\textbf{#1}} &
\multicolumn{3}{c}{\textbf{Temperature Prediction ($^\circ$C)}} &
\multirow{2}{*}{\textbf{Time (s)}} \\
\cmidrule(lr){2-4}
&
\textbf{RMSE $\downarrow$} & \textbf{\(R^2\) $\uparrow$} & \textbf{MAE $\downarrow$} &
\\}

\newcommand{\nnheadersplit}[1]{%
\multirow{2}{*}{\textbf{#1}} &
\multicolumn{3}{c}{\textbf{Temperature Prediction ($^\circ$C)}} &
\multicolumn{3}{c}{\textbf{Time (s)}} \\
\cmidrule(lr){2-4}
\cmidrule(lr){5-7}
&
\textbf{RMSE $\downarrow$} & \textbf{\(R^2\) $\uparrow$} & \textbf{MAE $\downarrow$} &
\textbf{Total} & \textbf{ECM} & \textbf{Prediction} \\}

\section{Results and discussion}
\label{sec:results}

This section first presents the main comparison between the proposed method and the measured-input baselines on both datasets (Table~\ref{tab:main_comparison}). With LoRA online adaptation, the proposed method gives the lowest RMSE at all three horizons on the first dataset and at the two shorter horizons on the second dataset; its advantage over all four learned baselines is statistically significant at all three horizons on the first dataset and at the one-step horizon on the second. Sections~\ref{sec:perf_first} and~\ref{sec:perf_second} discuss these results for each dataset. Section~\ref{sec:first_dataset} then explains, through ablation studies on the first dataset, how the final configuration was obtained. Section~\ref{sec:robustness} examines generalisation to the second dataset, transfer between datasets, sensitivity to random seeds and the original sampling interval, and Section~\ref{sec:exp_discussion} discusses the findings and limitations. Detailed results of the supporting experiments are given in Appendix~\ref{app:ablations}.

\begin{table}[H]
\centering
\caption{Main comparison with the measured-input baselines on both datasets (temperature errors in $^\circ$C). Boldface denotes the best value in each column; $\star$: significantly better than all four learned baselines after Holm correction. Standard deviations, $R^2$ and test times are given in Tables~\ref{tab:baselines_lco} and~\ref{tab:baselines_drive_cycle}.}
\label{tab:main_comparison}
\resizebox{\textwidth}{!}{%
\begin{tabular}{l cccccc cccccc}
\toprule
\multirow{3}{*}{\textbf{Method}} &
\multicolumn{6}{c}{\textbf{First dataset}} &
\multicolumn{6}{c}{\textbf{Second dataset}} \\
\cmidrule(lr){2-7}\cmidrule(lr){8-13}
& \multicolumn{2}{c}{\textbf{$h=1$ (10\,s)}} & \multicolumn{2}{c}{\textbf{$h=10$ (100\,s)}} & \multicolumn{2}{c}{\textbf{$h=30$ (300\,s)}}
& \multicolumn{2}{c}{\textbf{$h=1$ (10\,s)}} & \multicolumn{2}{c}{\textbf{$h=10$ (100\,s)}} & \multicolumn{2}{c}{\textbf{$h=30$ (300\,s)}} \\
\cmidrule(lr){2-3}\cmidrule(lr){4-5}\cmidrule(lr){6-7}\cmidrule(lr){8-9}\cmidrule(lr){10-11}\cmidrule(lr){12-13}
& \textbf{RMSE $\downarrow$} & \textbf{MAE $\downarrow$} & \textbf{RMSE $\downarrow$} & \textbf{MAE $\downarrow$} & \textbf{RMSE $\downarrow$} & \textbf{MAE $\downarrow$}
& \textbf{RMSE $\downarrow$} & \textbf{MAE $\downarrow$} & \textbf{RMSE $\downarrow$} & \textbf{MAE $\downarrow$} & \textbf{RMSE $\downarrow$} & \textbf{MAE $\downarrow$} \\
\midrule
Persistence & 0.0910 & -- & 0.7122 & -- & 1.8637 & -- & 0.0183 & -- & 0.1306 & -- & 0.2714 & -- \\
\midrule
XGBoost & 0.0655 & 0.0402 & 0.3486 & 0.1378 & 0.9405 & 0.3301 & 0.0074 & 0.0052 & 0.1134 & 0.0745 & 0.2509 & 0.1572 \\
LSTM & 0.0559 & 0.0369 & 0.2315 & 0.1161 & 0.7614 & 0.3018 & 0.0090 & 0.0065 & 0.1491 & 0.0985 & \textbf{0.2442} & 0.1448 \\
GRU & 0.0516 & 0.0344 & 0.2071 & 0.1076 & 0.7418 & 0.2948 & 0.0089 & 0.0065 & 0.1479 & 0.1001 & 0.2452 & \textbf{0.1427} \\
Transformer & 0.0513 & 0.0340 & 0.1910 & 0.1009 & 0.6730 & 0.2703 & 0.0087 & 0.0062 & 0.1100 & 0.0731 & 0.2583 & 0.1519 \\
\midrule
Proposed & 0.0499$^\star$ & 0.0334 & 0.1951 & 0.1001 & 0.6380 & 0.2546 & 0.0070$^\star$ & 0.0049 & 0.1163 & 0.0785 & 0.2650 & 0.1596 \\
Proposed + LoRA & \textbf{0.0491}$^\star$ & \textbf{0.0329} & \textbf{0.1703}$^\star$ & \textbf{0.0885} & \textbf{0.5867}$^\star$ & \textbf{0.2297} & \textbf{0.0065}$^\star$ & \textbf{0.0045} & \textbf{0.1037} & \textbf{0.0679} & 0.2556 & 0.1484 \\
\bottomrule
\end{tabular}%
}
\end{table}

\subsection{Performance on the first dataset}
\label{sec:perf_first}
Table~\ref{tab:main_comparison} summarises the results for the 82 test cycles of the first dataset; standard deviations, $R^2$ and test times are given in Table~\ref{tab:baselines_lco}. The Transformer is the strongest measured-input baseline by RMSE at all three horizons (0.0513, 0.1910 and 0.6730\,$^\circ$C). The proposed model without adaptation gives 0.0499, 0.1951 and 0.6380\,$^\circ$C; it significantly outperforms all four learned baselines at $h=1$, but its differences from the Transformer at $h=10$ and 30 are not significant ($p=0.072$ and $p=0.16$, respectively). With LoRA, RMSE decreases to 0.0491, 0.1703 and 0.5867\,$^\circ$C, with corresponding MAEs of 0.0329, 0.0885 and 0.2297\,$^\circ$C. The adapted configuration significantly outperforms all four learned baselines at each horizon, with significance retained after Holm correction over 24 comparisons. Persistence gives 0.0910, 0.7122 and 1.8637\,$^\circ$C, demonstrating that the forecasting error of an unchanged-temperature prediction increases substantially with the horizon.

For XGBoost, the ECM-enhanced configuration of the ablation study gives 0.0575, 0.2465 and 0.8060\,$^\circ$C (Table~\ref{tab:horizon_ablation}), compared with 0.0655, 0.3486 and 0.9405\,$^\circ$C for the measured-input baseline. These configurations also differ in target definition; the same-target feature ablation (Table~\ref{tab:feature_ablation}) provides the more direct one-step assessment of the ECM inputs. The proposed model requires 5.20--5.28\,s for the full test set, increasing to 81.59--85.31\,s with LoRA; the timing conventions differ from those of the measured-input baselines as described in Section~\ref{sec:experiment_implementation}.

\subsection{Performance on the second dataset}
\label{sec:perf_second}
The same comparison is conducted on the 25 test profiles of the second dataset (Table~\ref{tab:main_comparison}; details in Table~\ref{tab:baselines_drive_cycle}). At $h=1$, both the proposed model and its adapted variant significantly outperform all four learned baselines after Holm correction, with RMSEs of 0.0070 and 0.0065\,$^\circ$C, respectively, compared with 0.0074\,$^\circ$C for XGBoost, the strongest baseline. At $h=10$, the adapted model gives the lowest point-estimate RMSE (0.1037\,$^\circ$C). It significantly outperforms LSTM and GRU, but not XGBoost or the Transformer; the bootstrap CI for these latter comparisons includes zero. The unadapted model gives 0.1163\,$^\circ$C, compared with 0.1100\,$^\circ$C for the Transformer, and improves on the Transformer in only 6 of 25 profiles.

At $h=30$, LSTM gives the lowest RMSE (0.2442\,$^\circ$C) and GRU the lowest MAE (0.1427\,$^\circ$C). The adapted model gives RMSE 0.2556\,$^\circ$C, with no significant improvement over the learned baselines; the unadapted model is significantly worse than Transformer, GRU and LSTM. The ECM-enhanced XGBoost results in Table~\ref{tab:drive_cycle_horizon} are also numerically worse than those of measured-input XGBoost at all three horizons. As on the first dataset, the two XGBoost experiments differ in target definition. Therefore, the results do not isolate the effect of ECM inputs on the second dataset. On the second dataset, a statistically significant improvement over all measured-input baselines is established only at the one-step horizon.

\subsection{Ablation studies}
\label{sec:first_dataset}
This subsection explains how the configuration evaluated in Sections~\ref{sec:perf_first} and~\ref{sec:perf_second} was obtained. The experiments progressively compare ECM configurations, temperature-model inputs, temporal and channel architectures, online adaptation and the prediction target. The input-feature, regression-model, architecture and online-adaptation comparisons are reported in full here; the ECM, lookback, horizon, hidden-state and target ablations are summarised, with their tables in Appendix~\ref{app:ablations}. Because the validation cell LCO\_4 is used for early stopping, these configuration choices are based on the reported test-cell results of the first dataset and may therefore be optimistic for this dataset. The second dataset, on which no configuration is tuned, provides an independent evaluation of the selected configuration.

\subsubsection{Equivalent circuit model}
The ECM configuration is selected with XGBoost and a 10-step lookback; the five ECM ablations are reported in Appendix~\ref{app:ecm} (Tables~\ref{tab:ocv_soc_ablation}--\ref{tab:filter_ablation}). The exponential OCV--SOC mapping gives the lowest temperature RMSE (0.0587\,$^\circ$C) among the five evaluated mappings. Two RC networks are retained, since a third branch converges to the same parameters as the second. One-state hysteresis~\cite{plett2004ekf2} reduces temperature RMSE to 0.0578\,$^\circ$C, although voltage RMSE increases from 77.98 to 79.20\,mV, whereas temperature-dependent (Arrhenius) resistances increase temperature RMSE to 0.0591\,$^\circ$C. Among the KF, EKF, UKF and particle filter, the EKF gives the lowest voltage, SOC and temperature RMSE (71.38\,mV, 4.33\,\% and 0.0574\,$^\circ$C). The selected ECM therefore comprises two RC networks, exponential OCV, hysteresis and EKF-based state estimation, without explicit thermal coupling.

\subsubsection{Input features}
Table~\ref{tab:feature_ablation} compares windows of measured current, voltage and temperature ($I$, $V$ and $T$) with features derived from the selected ECM. Adding measured temperature to the $I,V$ window reduces RMSE from 0.0683 to 0.0619\,$^\circ$C (9.4\,\%) and MAE from 0.0448 to 0.0394\,$^\circ$C (12.1\,\%). Adding SOC at the current step or over the window further reduces RMSE to 0.0595 or 0.0589\,$^\circ$C, respectively. The lowest RMSE (0.0574\,$^\circ$C) is obtained by combining the measurement window with the current ECM states $[\mathrm{SOC},U_1,U_2,H]$. Using the full ECM-state window instead gives 0.0578\,$^\circ$C, while using only the ECM-state window gives 0.0620\,$^\circ$C. These comparisons support using the measured history together with the current ECM context on this dataset. The inference time includes ECM estimation except for the measured-input-only variants.

\begin{table}[H]
\centering
\caption{Ablation of measured and ECM-derived input features.}
\label{tab:feature_ablation}

\resizebox{\textwidth}{!}{%
\begin{tabular}{lcccc}
\toprule
\nnheader{Input Features}
\midrule

\multicolumn{5}{l}{\textbf{Raw Measurement Features}} \\
\midrule

I, V, T (window) &
0.0619 $\pm$ 0.0005 & 1.0000 $\pm$ 0.0000 & 0.0394 $\pm$ 0.0002 &
\textbf{0.10} \\

I, V (window) &
0.0683 $\pm$ 0.0004 & 1.0000 $\pm$ 0.0000 & 0.0448 $\pm$ 0.0002 &
0.11 \\

\midrule
\multicolumn{5}{l}{\textbf{SOC-Enhanced Features}} \\
\midrule

I, V, T (window) + SOC (current step) &
0.0595 $\pm$ 0.0004 & 1.0000 $\pm$ 0.0000 & 0.0389 $\pm$ 0.0002 &
8.26 \\

I, V, T (window) + SOC (window) &
0.0589 $\pm$ 0.0003 & 1.0000 $\pm$ 0.0000 & 0.0386 $\pm$ 0.0002 &
8.18 \\

\midrule
\multicolumn{5}{l}{\textbf{ECM State-Enhanced Features}} \\
\midrule

I, V, T (window) + [SOC, $U_1$, $U_2$, $H$] (current step) &
\textbf{0.0574 $\pm$ 0.0003} & 1.0000 $\pm$ 0.0000 & \textbf{0.0378 $\pm$ 0.0002} &
8.21 \\

I, V, T (window) + [SOC, $U_1$, $U_2$, $H$] (window) &
0.0578 $\pm$ 0.0003 & 1.0000 $\pm$ 0.0000 & 0.0382 $\pm$ 0.0002 &
8.26 \\

\midrule
\multicolumn{5}{l}{\textbf{Physics-State-Only Features}} \\
\midrule

[SOC, $U_1$, $U_2$, $H$] only (window) &
0.0620 $\pm$ 0.0004 & 1.0000 $\pm$ 0.0000 & 0.0402 $\pm$ 0.0002 &
8.20 \\

\bottomrule
\end{tabular}%
}
\end{table}

\subsubsection{Lookback window and prediction horizon}
Longer XGBoost lookbacks reduce temperature RMSE, from 0.0574\,$^\circ$C at 10 steps to 0.0525\,$^\circ$C at 200 steps, but part of this gain results from excluding the beginning of each cycle: on common test points ($t\geq2000$\,s), lookbacks of 1, 5, 50 and 200 steps give 0.0558, 0.0557, 0.0530 and 0.0525\,$^\circ$C, respectively (Appendix~\ref{app:lookback}, Table~\ref{tab:lookback_ablation}). A lookback of 10 steps is retained to avoid requiring long measurement histories. With this lookback, XGBoost RMSE increases from 0.0575\,$^\circ$C at 10\,s to 0.2465\,$^\circ$C at 100\,s and 1.4043\,$^\circ$C at 600\,s (Table~\ref{tab:horizon_ablation}). The architecture ablations below use the one-step horizon.

\subsubsection{Temperature regression model}
The regression comparison in Table~\ref{tab:model_ablation} uses the measured window and current ECM context. For LSTM, GRU and Transformer, the context is concatenated to the sequence-model output before the dense layer. The Transformer achieves the lowest temperature RMSE and MAE (0.0515 and 0.0338\,$^\circ$C), compared with RMSEs of 0.0559 for GRU, 0.0562 for MLP, 0.0564 for LSTM and 0.0575\,$^\circ$C for XGBoost. It is consequently selected as the temporal handler for the next comparison.

\begin{table}[H]
\centering
\caption{Temperature-regression model comparison using the selected inputs.}
\label{tab:model_ablation}

\resizebox{\textwidth}{!}{%
\begin{tabular}{lcccccc}
\toprule
\nnheadersplit{Regression Model}
\midrule

XGBoost &
0.0575 $\pm$ 0.0003 & 1.0000 $\pm$ 0.0000 & 0.0377 $\pm$ 0.0002 &
5.17 & 5.12 & 0.05 \\

ElasticNet &
0.0671 $\pm$ 0.0004 & 1.0000 $\pm$ 0.0000 & 0.0443 $\pm$ 0.0002 &
5.12 & 5.11 & \textbf{$<$0.01} \\

Random Forest &
0.0593 $\pm$ 0.0003 & 1.0000 $\pm$ 0.0000 & 0.0400 $\pm$ 0.0002 &
5.39 & 5.21 & 0.17 \\

MLP &
0.0562 $\pm$ 0.0003 & 1.0000 $\pm$ 0.0000 & 0.0371 $\pm$ 0.0002 &
5.10 & 5.08 & 0.02 \\

LSTM &
0.0564 $\pm$ 0.0003 & 1.0000 $\pm$ 0.0000 & 0.0372 $\pm$ 0.0002 &
5.16 & 4.97 & 0.19 \\

GRU &
0.0559 $\pm$ 0.0003 & 1.0000 $\pm$ 0.0000 & 0.0368 $\pm$ 0.0002 &
5.23 & 5.07 & 0.16 \\

Transformer &
\textbf{0.0515 $\pm$ 0.0003} & 1.0000 $\pm$ 0.0000 & \textbf{0.0338 $\pm$ 0.0002} &
5.39 & 5.20 & 0.19 \\

\bottomrule
\end{tabular}%
}
\end{table}

\subsubsection{Temporal and channel handlers}
Table~\ref{tab:temporal_channel_ablation} compares the temporal Transformer with channel-oriented models and a combined architecture. The temporal Transformer receives ECM states either at each input step or as current-step context after the sequence encoder. The channel handlers take $I,V,T$ as channels and concatenate current ECM context before the dense layer. In the iTransformer, each channel's lookback series forms one token and attention operates across channels; TSMixer alternates time-mixing and channel-mixing MLPs. The combined architecture runs the temporal Transformer with ECM states at its input and the iTransformer in parallel and concatenates their outputs with the context.

The iTransformer achieves the lowest temperature RMSE (0.0510\,$^\circ$C), compared with 0.0514 and 0.0515\,$^\circ$C for the two temporal variants, 0.0570\,$^\circ$C for TSMixer and 0.0521\,$^\circ$C for the combined architecture. It also has the lowest prediction time in this comparison (0.05\,s). The iTransformer is therefore retained as the base temperature model; the evaluated parallel temporal--channel combination provides no additional accuracy benefit.

\begin{table}[H]
\centering
\caption{Comparison of temporal, channel and combined handlers.}
\label{tab:temporal_channel_ablation}

\resizebox{\textwidth}{!}{%
\begin{tabular}{lcccccc}
\toprule
\nnheadersplit{Configuration}
\midrule

\multicolumn{7}{l}{\textbf{Temporal handler (Transformer)}} \\
\midrule
ECM states at input &
0.0514 $\pm$ 0.0003 & 1.0000 $\pm$ 0.0000 & 0.0339 $\pm$ 0.0002 &
5.20 & 4.98 & 0.22 \\
ECM states at hidden output &
0.0515 $\pm$ 0.0003 & 1.0000 $\pm$ 0.0000 & 0.0338 $\pm$ 0.0002 &
5.39 & 5.20 & 0.19 \\

\midrule
\multicolumn{7}{l}{\textbf{Channel handler (channels: I, V, T; context: SOC, $U_1$, $U_2$, $H$)}} \\
\midrule
iTransformer &
\textbf{0.0510 $\pm$ 0.0003} & 1.0000 $\pm$ 0.0000 & \textbf{0.0336 $\pm$ 0.0002} &
5.08 & 5.03 & \textbf{0.05} \\
TSMixer &
0.0570 $\pm$ 0.0003 & 1.0000 $\pm$ 0.0000 & 0.0377 $\pm$ 0.0002 &
5.13 & 5.03 & 0.11 \\

\midrule
\multicolumn{7}{l}{\textbf{Temporal--channel handler}} \\
\midrule
Transformer (ECM states at input) + iTransformer &
0.0521 $\pm$ 0.0003 & 1.0000 $\pm$ 0.0000 & 0.0350 $\pm$ 0.0002 &
5.37 & 5.11 & 0.26 \\

\bottomrule
\end{tabular}%
}
\end{table}

\subsubsection{Recurrent hidden state}
The iTransformer is extended with a 64-unit hidden state that is passed between consecutive steps and updated either by a linear layer with ReLU or by GRU gating (Appendix~\ref{app:hidden}, Table~\ref{tab:hidden_state_ablation}). At lookback 10, GRU gating reduces RMSE from 0.0510 to 0.0498\,$^\circ$C, whereas the linear update increases it to 0.0550\,$^\circ$C; GRU gating also improves RMSE at lookback 1, from 0.0571 to 0.0555\,$^\circ$C. Prediction time increases from 0.05 to 0.22\,s. The GRU-gated model with lookback 10 is retained for the continual-learning experiments.

\subsubsection{Continual learning on new cells}
\label{sec:continual_experiments}
For online adaptation, each test cell starts independently from the trained model. Its cycles are processed in chronological order in blocks of 64 steps (640\,s). At the one-step horizon, each block is first predicted and its predictions retained for evaluation; the measured target temperatures are subsequently used for one update. The model is updated every 64 steps, using only samples whose target time $t+h$ is already observed ($t+h\leq k$, where $k$ is the current step). For longer horizons, this corresponds to an additional delay of $h-1$ steps. Remaining eligible samples are used at the end of a cycle. Thus, no future information is used.

Table~\ref{tab:continual_ablation} compares adaptation of the recurrent iTransformer with and without a key--value memory layer. The memory-augmented variant uses 4096 slots, four heads and the top 32 keys, with output $y_t=\mathrm{Linear}(h_t+\mathrm{Memory}(h_t))$. Sparse memory fine-tuning~\cite{lin2025sparse} uses SGD to update only the values of the top 64 memory slots ranked by TF--IDF of their accesses in the block; IDF is obtained from the training cells. Alternatives update all accessed slots or rank slots by term frequency (TF) alone. Full fine-tuning uses AdamW with learning rate $10^{-4}$ and weight decay 0.1. For low-rank adaptation (LoRA), adapters of rank $r=8$, with $\alpha=16$ and scaling $\alpha/r$, are added to every linear layer and to the GRU cell input-to-hidden and hidden-to-hidden weights. The LoRA matrix $B$ is initialised to zero, so the adapted model starts identical to the trained model. The original weights are frozen; only the LoRA parameters are updated, using AdamW with learning rate $10^{-3}$ and weight decay 0.1. The default sparse-memory configuration updates every 64 steps with learning rate 1.

Without adaptation, adding the memory layer increases temperature RMSE from 0.0498 to 0.0503\,$^\circ$C. Full fine-tuning of the memory-augmented model reduces RMSE to 0.0495\,$^\circ$C, and LoRA gives 0.0493\,$^\circ$C. Updating all accessed slots, TF-selected slots or TF--IDF-selected slots gives 0.0501, 0.0501 and 0.0502\,$^\circ$C, respectively. Updating the TF--IDF-selected slots every step improves RMSE to 0.0497\,$^\circ$C but increases total test time from 14.33 to 182.26\,s. Increasing the learning rate to 5 gives 0.0501\,$^\circ$C, whereas increasing it to 20 worsens RMSE to 0.0526\,$^\circ$C and increases validation MSE from 0.3709 to 0.9430.

LoRA applied to the model without memory gives the same reported RMSE (0.0493\,$^\circ$C) as LoRA applied to the memory model, and significantly improves on the unadapted memory-free configuration. The validation MSE decreases from 0.3748 to 0.3699. Total test time increases from 5.34 to 85.58\,s, with prediction and updates requiring 1.27\,ms per step. The memory-free LoRA configuration is retained, since it matches the lowest reported RMSE without an additional memory layer. In Table~\ref{tab:continual_ablation}, $\dagger$ denotes a significant improvement over the unadapted memory-free model and $\ddagger$ a significant improvement over the unadapted memory-augmented model, using the joint CI and Wilcoxon criteria.

\begin{table}[H]
\centering
\caption{Online adaptation on the first dataset ($h=1$). $\dagger$: improvement over the memory-free reference; $\ddagger$: improvement over the memory-augmented reference (significant).}
\label{tab:continual_ablation}

\resizebox{\textwidth}{!}{%
\begin{tabular}{lcccccccc}
\toprule

\multirow{2}{*}{\textbf{Method}} &
\multicolumn{3}{c}{\textbf{Temperature Prediction ($^\circ$C)}} &
\multirow{2}{*}{\textbf{\makecell{LCO\_4 MSE\\before $\rightarrow$ after}}} &
\multicolumn{4}{c}{\textbf{Time}} \\
\cmidrule(lr){2-4}
\cmidrule(lr){6-9}&
\textbf{RMSE $\downarrow$} & \textbf{\(R^2\) $\uparrow$} & \textbf{MAE $\downarrow$} & &
\textbf{Total (s)} & \textbf{ECM (s)} & \textbf{Prediction (s)} & \textbf{\makecell{Prediction\\per step (ms)}} \\
\midrule

\multicolumn{9}{l}{\textbf{No adaptation}} \\
\midrule
Without memory (reference) &
0.0498 $\pm$ 0.0003 & 1.0000 $\pm$ 0.0000 & 0.0331 $\pm$ 0.0002 & 0.3748 &
5.34 & 5.11 & \textbf{0.22} & 0.003 \\
+ memory layer &
0.0503 $\pm$ 0.0003 & 1.0000 $\pm$ 0.0000 & 0.0330 $\pm$ 0.0002 & 0.3709 &
5.69 & 5.03 & 0.66 & 0.010 \\

\midrule
\multicolumn{9}{l}{\textbf{Online adaptation of the memory model}} \\
\midrule
Full finetuning &
0.0495 $\pm$ 0.0003$^\ddagger$ & 1.0000 $\pm$ 0.0000 & 0.0327 $\pm$ 0.0002 & 0.3709 $\rightarrow$ \textbf{0.3655} &
46.07 & 5.11 & 40.96 & 0.65 \\
LoRA &
0.0493 $\pm$ 0.0003$^{\dagger\ddagger}$ & 1.0000 $\pm$ 0.0000 & \textbf{0.0325 $\pm$ 0.0002} & 0.3709 $\rightarrow$ 0.3659 &
86.89 & 5.22 & 81.67 & 1.29 \\
Memory, all accessed slots &
0.0501 $\pm$ 0.0003$^\ddagger$ & 1.0000 $\pm$ 0.0000 & 0.0329 $\pm$ 0.0002 & 0.3709 $\rightarrow$ 0.3731 &
13.94 & 5.05 & 8.89 & 0.14 \\
Memory, top-$t$ by TF &
0.0501 $\pm$ 0.0003$^\ddagger$ & 1.0000 $\pm$ 0.0000 & 0.0330 $\pm$ 0.0002 & 0.3709 $\rightarrow$ 0.3750 &
14.25 & 5.05 & 9.19 & 0.14 \\
Sparse memory, top-$t$ by TF--IDF &
0.0502 $\pm$ 0.0003$^\ddagger$ & 1.0000 $\pm$ 0.0000 & 0.0329 $\pm$ 0.0002 & 0.3709 $\rightarrow$ 0.3701 &
14.33 & 5.07 & 9.27 & 0.15 \\

\midrule
\multicolumn{9}{l}{\textbf{Sparse memory finetuning: update frequency and learning rate (default: every 64 steps, learning rate 1)}} \\
\midrule
Update every step &
0.0497 $\pm$ 0.0003$^\ddagger$ & 1.0000 $\pm$ 0.0000 & 0.0326 $\pm$ 0.0002 & 0.3709 $\rightarrow$ 0.3710 &
182.26 & 5.06 & 177.21 & 2.79 \\
Learning rate 5 &
0.0501 $\pm$ 0.0003$^\ddagger$ & 1.0000 $\pm$ 0.0000 & 0.0329 $\pm$ 0.0002 & 0.3709 $\rightarrow$ 0.3718 &
14.31 & 5.03 & 9.28 & 0.15 \\
Learning rate 20 &
0.0526 $\pm$ 0.0003 & 1.0000 $\pm$ 0.0000 & 0.0337 $\pm$ 0.0002 & 0.3709 $\rightarrow$ 0.9430 &
14.28 & 5.04 & 9.24 & 0.15 \\

\midrule
\multicolumn{9}{l}{\textbf{Online adaptation of the model without memory}} \\
\midrule
LoRA &
\textbf{0.0493 $\pm$ 0.0003}$^\dagger$ & 1.0000 $\pm$ 0.0000 & 0.0329 $\pm$ 0.0001 & 0.3748 $\rightarrow$ 0.3699 &
85.58 & 4.94 & 80.64 & 1.27 \\

\bottomrule
\end{tabular}%
}
\end{table}

\begin{figure}[H]
\centering
\includegraphics[width=0.9\textwidth]{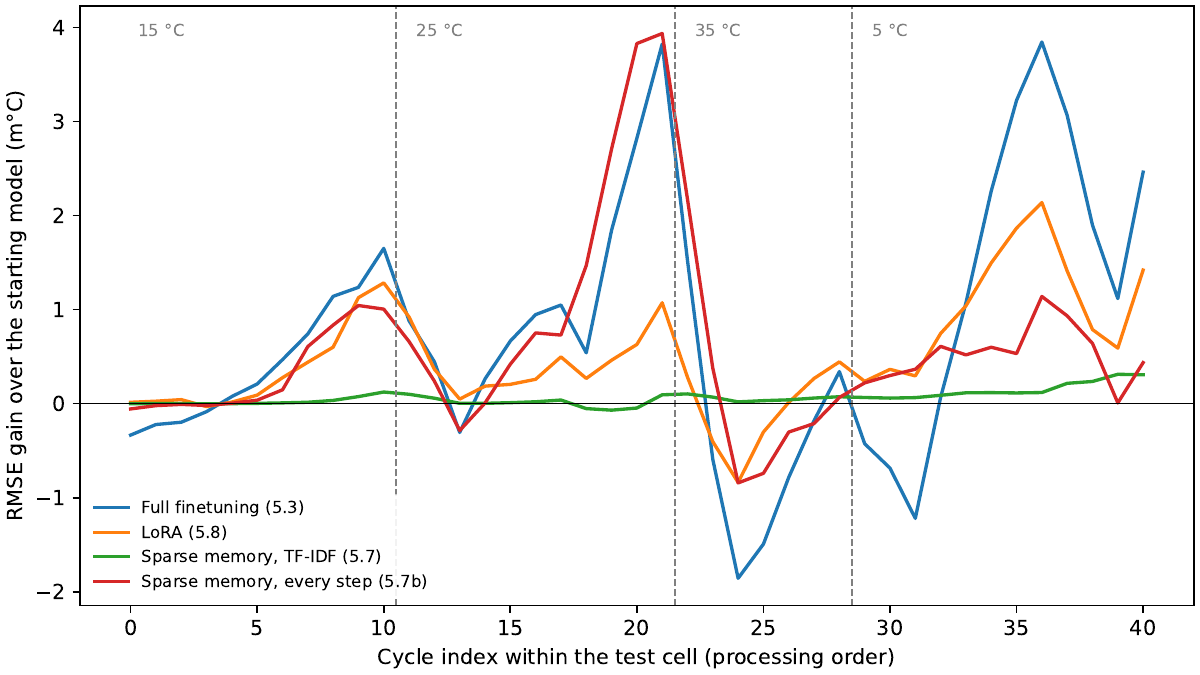}
\caption{Cycle-wise RMSE reduction from online adaptation, shown as a three-cycle moving average.}
\label{fig:continual_curve}
\end{figure}

Figure~\ref{fig:continual_curve} presents the per-cycle reduction in RMSE relative to the corresponding starting model, averaged over LCO\_5 and LCO\_6 using a three-cycle moving average. The reference is the memory-augmented model for full and sparse memory fine-tuning and the memory-free model for LoRA. Cycles are processed in the order of ambient temperatures 15, 25, 35 and 5\,$^\circ$C; dashed lines identify changes in ambient temperature. A positive value denotes an improvement, and adaptation prior to a given prediction is restricted to the measured targets of earlier cycles or eligible earlier blocks of the current cycle.

\subsubsection{Prediction horizon with online adaptation}
At horizons from 10 to 600\,s (Appendix~\ref{app:horizon_adapt}, Table~\ref{tab:horizon_adapt_ablation} and Fig.~\ref{fig:horizon}), LoRA reduces RMSE relative to the unadapted model at every horizon, for example from 0.2003 to 0.1752\,$^\circ$C at 100\,s and from 0.7060 to 0.6398\,$^\circ$C at 300\,s. The reductions are significant for $h\leq30$, where LoRA also significantly outperforms XGBoost. However, validation MSE increases after adaptation at several longer horizons, for example from 0.2539 to 0.5022 at $h=30$.

\subsubsection{Physics-informed inputs and prediction target}
Adding a heat-generation feature, the ambient-temperature difference or a lumped thermal residual model does not significantly improve the recurrent iTransformer (Appendix~\ref{app:target}, Table~\ref{tab:physics_ablation}). Among the prediction targets (Table~\ref{tab:target_comparison}), direct prediction of $T_{t+h}$ is significantly worse than the temperature change $\Delta T=T_{t+h}-T_t$ at all three horizons after Holm correction, and the relative change $(T_{t+h}-T_t)/T_t$ is significantly worse than $\Delta T$ without adaptation at $h=10$ and 30. With $\Delta T$, the RMSEs are 0.0499, 0.1951 and 0.6380\,$^\circ$C without adaptation and 0.0491, 0.1703 and 0.5867\,$^\circ$C with LoRA at $h=1$, 10 and 30, respectively. The final configuration therefore uses the electrical 2-RC ECM with exponential OCV, hysteresis and EKF, the iTransformer with GRU hidden state, a 10-step measured-input window and the $\Delta T$ target, with LoRA as the online-adaptation variant.

\subsection{Generalisation and robustness}
\label{sec:robustness}
\subsubsection{Application to the second dataset}
\label{sec:second_dataset}
The selected ECM settings are refitted on the training profiles of the second dataset (Appendix~\ref{app:second}, Table~\ref{tab:drive_cycle_ecm}). Voltage, SOC and one-step temperature RMSEs are lower than on the first dataset (57.93 versus 71.38\,mV, 1.12 versus 4.33\,\% and 0.0076 versus 0.0575\,$^\circ$C), consistent with the relatively slow temperature changes (1--7\,$^\circ$C per profile) of these profiles; the errors are therefore not directly comparable across datasets. With the selected architecture (Table~\ref{tab:drive_cycle_horizon}), LoRA significantly improves on the unadapted model at each horizon, reducing RMSE from 0.0070 to 0.0065, from 0.1163 to 0.1037 and from 0.2650 to 0.2556\,$^\circ$C, although validation MSE increases after adaptation at all three horizons.

\subsubsection{Transfer learning between datasets}
The transfer study in Table~\ref{tab:transfer} assesses weight reuse in both directions, using the seed-42 model trained on the source dataset at the corresponding horizon. All configurations use lookback 10 and the $\Delta T$ target, without online adaptation. Zero-shot transfer involves no training; head-only fine-tuning freezes the iTransformer and GRU and trains the 65 output-layer parameters, whereas full fine-tuning updates all 40\,833 parameters with learning rate $10^{-4}$. Fine-tuning uses the target training data and early stopping on the target validation data, with rescaling based on the target training data. From-scratch models are trained on the target dataset only. In all transfer settings, ECM states are computed with the ECM fitted on the target dataset. Zero-shot transfer applies the source model with the scalers of the source training data. Head-only fine-tuning uses learning rate $10^{-3}$; both fine-tuning variants use Adam and the training schedule of the recurrent models (at most 300 epochs, minimum 100 epochs, patience 30).

Zero-shot transfer is significantly worse than training from scratch in both directions at every horizon. In the LCO-to-NMC direction, full fine-tuning at $h=10$ and head-only fine-tuning at $h=30$ yield slightly lower point estimates than training from scratch (0.1151 versus 0.1163\,$^\circ$C and 0.2572 versus 0.2650\,$^\circ$C, respectively), but these differences are not significant. At $h=1$, both fine-tuning variants are significantly worse. In the NMC-to-LCO direction, all transfer variants are significantly worse than training from scratch at every horizon, including full fine-tuning. No transfer variant significantly improves on training from scratch after Holm correction over nine comparisons per direction. The results therefore support reusing the selected configuration across these datasets more clearly than reusing its learned weights.

\begin{table}[H]
\centering
\caption{Bidirectional transfer learning without online adaptation. $\dagger$: significantly worse than training from scratch after Holm correction.}
\label{tab:transfer}

\resizebox{\textwidth}{!}{%
\begin{tabular}{cl c ccc}
\toprule
\multirow{2}{*}{\textbf{\makecell{Horizon\\(steps)}}} &
\multirow{2}{*}{\textbf{Method}} &
\multirow{2}{*}{\textbf{\makecell{Trainable\\parameters}}} &
\multicolumn{3}{c}{\textbf{Temperature Prediction ($^\circ$C)}} \\
\cmidrule(lr){4-6}
& & & \textbf{RMSE $\downarrow$} & \textbf{\(R^2\) $\uparrow$} & \textbf{MAE $\downarrow$} \\
\midrule
\multicolumn{6}{l}{\textit{(a) Dataset 1 $\rightarrow$ Dataset 2 (evaluated on the second dataset, 25 test profiles)}} \\
\midrule
\multirow{4}{*}{1 (10\,s)}
& From scratch (target only) & 40{,}833 & \textbf{0.0070 $\pm$ 0.0001} & 1.0000 $\pm$ 0.0000 & 0.0049 $\pm$ 0.0000 \\
& Zero-shot & 0 & 0.1012 $\pm$ 0.0004$^\dagger$ & 0.9999 $\pm$ 0.0000 & 0.0785 $\pm$ 0.0003 \\
& Fine-tune head & 65 & 0.0126 $\pm$ 0.0002$^\dagger$ & 1.0000 $\pm$ 0.0000 & 0.0085 $\pm$ 0.0000 \\
& Fine-tune all & 40{,}833 & 0.0077 $\pm$ 0.0001$^\dagger$ & 1.0000 $\pm$ 0.0000 & 0.0055 $\pm$ 0.0000 \\
\midrule
\multirow{4}{*}{10 (100\,s)}
& From scratch (target only) & 40{,}833 & 0.1163 $\pm$ 0.0010 & 0.9999 $\pm$ 0.0000 & 0.0785 $\pm$ 0.0005 \\
& Zero-shot & 0 & 0.6085 $\pm$ 0.0029$^\dagger$ & 0.9980 $\pm$ 0.0000 & 0.4642 $\pm$ 0.0022 \\
& Fine-tune head & 65 & 0.1155 $\pm$ 0.0011 & 0.9999 $\pm$ 0.0000 & 0.0761 $\pm$ 0.0005 \\
& Fine-tune all & 40{,}833 & \textbf{0.1151 $\pm$ 0.0010} & 0.9999 $\pm$ 0.0000 & 0.0778 $\pm$ 0.0005 \\
\midrule
\multirow{4}{*}{30 (300\,s)}
& From scratch (target only) & 40{,}833 & 0.2650 $\pm$ 0.0036 & 0.9996 $\pm$ 0.0000 & 0.1596 $\pm$ 0.0012 \\
& Zero-shot & 0 & 1.7684 $\pm$ 0.0072$^\dagger$ & 0.9834 $\pm$ 0.0002 & 1.4034 $\pm$ 0.0058 \\
& Fine-tune head & 65 & \textbf{0.2572 $\pm$ 0.0036} & 0.9996 $\pm$ 0.0000 & 0.1542 $\pm$ 0.0011 \\
& Fine-tune all & 40{,}833 & 0.2892 $\pm$ 0.0027 & 0.9996 $\pm$ 0.0000 & 0.2008 $\pm$ 0.0011 \\
\midrule
\multicolumn{6}{l}{\textit{(b) Dataset 2 $\rightarrow$ Dataset 1 (evaluated on the first dataset, 82 test cycles)}} \\
\midrule
\multirow{4}{*}{1 (10\,s)}
& From scratch (target only) & 40{,}833 & \textbf{0.0499 $\pm$ 0.0003} & 1.0000 $\pm$ 0.0000 & 0.0334 $\pm$ 0.0001 \\
& Zero-shot & 0 & 0.0779 $\pm$ 0.0005$^\dagger$ & 0.9999 $\pm$ 0.0000 & 0.0480 $\pm$ 0.0002 \\
& Fine-tune head & 65 & 0.0583 $\pm$ 0.0003$^\dagger$ & 1.0000 $\pm$ 0.0000 & 0.0384 $\pm$ 0.0002 \\
& Fine-tune all & 40{,}833 & 0.0552 $\pm$ 0.0003$^\dagger$ & 1.0000 $\pm$ 0.0000 & 0.0365 $\pm$ 0.0002 \\
\midrule
\multirow{4}{*}{10 (100\,s)}
& From scratch (target only) & 40{,}833 & \textbf{0.1951 $\pm$ 0.0044} & 0.9996 $\pm$ 0.0000 & 0.1001 $\pm$ 0.0007 \\
& Zero-shot & 0 & 0.6459 $\pm$ 0.0060$^\dagger$ & 0.9960 $\pm$ 0.0001 & 0.2874 $\pm$ 0.0022 \\
& Fine-tune head & 65 & 0.3396 $\pm$ 0.0042$^\dagger$ & 0.9989 $\pm$ 0.0000 & 0.1765 $\pm$ 0.0011 \\
& Fine-tune all & 40{,}833 & 0.2118 $\pm$ 0.0045$^\dagger$ & 0.9996 $\pm$ 0.0000 & 0.1106 $\pm$ 0.0007 \\
\midrule
\multirow{4}{*}{30 (300\,s)}
& From scratch (target only) & 40{,}833 & \textbf{0.6380 $\pm$ 0.0120} & 0.9961 $\pm$ 0.0001 & 0.2546 $\pm$ 0.0024 \\
& Zero-shot & 0 & 1.8183 $\pm$ 0.0161$^\dagger$ & 0.9680 $\pm$ 0.0006 & 0.7882 $\pm$ 0.0064 \\
& Fine-tune head & 65 & 1.2001 $\pm$ 0.0134$^\dagger$ & 0.9861 $\pm$ 0.0003 & 0.5899 $\pm$ 0.0041 \\
& Fine-tune all & 40{,}833 & 0.7269 $\pm$ 0.0132$^\dagger$ & 0.9949 $\pm$ 0.0002 & 0.3030 $\pm$ 0.0027 \\
\bottomrule
\end{tabular}%
}
\end{table}

\subsubsection{Sensitivity to random seeds}
Tables~\ref{tab:seeds_lco} and~\ref{tab:seeds_drive_cycle} report results for temperature-model seeds 42, 1 and 2. The ECM, partitions and other settings remain unchanged for the first-dataset seed study. On the first dataset, LoRA lowers RMSE for every seed and horizon, giving across-seed mean RMSEs of $0.0493\pm0.0003$, $0.1713\pm0.0010$ and $0.6018\pm0.0229$\,$^\circ$C. Even the largest adapted RMSE across the three seeds at each horizon is lower than that of the single-run Transformer baseline (0.0513, 0.1910 and 0.6730\,$^\circ$C, respectively).

\begin{table}[H]
\centering
\caption{First-dataset temperature RMSE ($^\circ$C) across three random seeds (mean and sample standard deviation).}
\label{tab:seeds_lco}
\resizebox{\textwidth}{!}{%
\begin{tabular}{cl cccc}
\toprule
\textbf{\makecell{Horizon\\(steps)}} & \textbf{Method} &
\textbf{Seed 42} & \textbf{Seed 1} & \textbf{Seed 2} & \textbf{Mean $\pm$ std} \\
\midrule
\multirow{2}{*}{1 (10\,s)}
& Proposed & 0.0499 & 0.0505 & 0.0501 & 0.0502 $\pm$ 0.0003 \\
& Proposed + LoRA & 0.0491 & 0.0496 & 0.0493 & \textbf{0.0493 $\pm$ 0.0003} \\
\midrule
\multirow{2}{*}{10 (100\,s)}
& Proposed & 0.1951 & 0.1893 & 0.1866 & 0.1903 $\pm$ 0.0043 \\
& Proposed + LoRA & 0.1703 & 0.1722 & 0.1713 & \textbf{0.1713 $\pm$ 0.0010} \\
\midrule
\multirow{2}{*}{30 (300\,s)}
& Proposed & 0.6380 & 0.6333 & 0.6711 & 0.6475 $\pm$ 0.0206 \\
& Proposed + LoRA & 0.5867 & 0.5905 & 0.6281 & \textbf{0.6018 $\pm$ 0.0229} \\
\bottomrule
\end{tabular}%
}
\end{table}

On the second dataset, LoRA likewise reduces RMSE for every seed and horizon, giving across-seed mean RMSEs of $0.0063\pm0.0002$, $0.1066\pm0.0037$ and $0.2546\pm0.0016$\,$^\circ$C. Compared with the strongest single-run baseline at each horizon, the adapted model achieves lower RMSE for all three seeds at $h=1$, for two seeds at $h=10$, and for none at $h=30$. The comparisons are consistent with the horizon-dependent baseline results, but do not quantify baseline sensitivity to seed because the baselines have only single-run results in the reported comparisons.

\begin{table}[H]
\centering
\caption{Second-dataset temperature RMSE ($^\circ$C) across three random seeds (mean and sample standard deviation).}
\label{tab:seeds_drive_cycle}
\resizebox{\textwidth}{!}{%
\begin{tabular}{cl cccc}
\toprule
\textbf{\makecell{Horizon\\(steps)}} & \textbf{Method} &
\textbf{Seed 42} & \textbf{Seed 1} & \textbf{Seed 2} & \textbf{Mean $\pm$ std} \\
\midrule
\multirow{2}{*}{1 (10\,s)}
& Proposed & 0.0070 & 0.0064 & 0.0063 & 0.0066 $\pm$ 0.0004 \\
& Proposed + LoRA & 0.0065 & 0.0063 & 0.0061 & \textbf{0.0063 $\pm$ 0.0002} \\
\midrule
\multirow{2}{*}{10 (100\,s)}
& Proposed & 0.1163 & 0.1357 & 0.1169 & 0.1230 $\pm$ 0.0110 \\
& Proposed + LoRA & 0.1037 & 0.1108 & 0.1054 & \textbf{0.1066 $\pm$ 0.0037} \\
\midrule
\multirow{2}{*}{30 (300\,s)}
& Proposed & 0.2650 & 0.2597 & 0.2615 & 0.2621 $\pm$ 0.0027 \\
& Proposed + LoRA & 0.2556 & 0.2527 & 0.2554 & \textbf{0.2546 $\pm$ 0.0016} \\
\bottomrule
\end{tabular}%
}
\end{table}

\subsubsection{Original sampling interval}
\label{sec:original_sampling}
The second dataset is additionally evaluated at its original sampling interval $\Delta t=0.1$\,s (Table~\ref{tab:dt01_drive_cycle}). The lookback and horizon lengths in steps remain unchanged: lookback 10 corresponds to 1\,s, while $h=1$, 10 and 30 correspond to 0.1, 1 and 3\,s. The same data partition, fitted ECM configuration and $\Delta T$ target are used; the measured-input baselines continue to use $I,V,T$ only. Neural models are trained for at most 30 epochs, with a minimum of 10 epochs and patience 5. Results are from single training runs with seed 42, and test times are medians of five inference runs. No confidence intervals or significance tests are reported. The ECM fitted on the drive-cycle training data at the 10\,s setting is reused without refitting. One 64-step online update block corresponds to 6.4\,s.

At $h=1$, all methods match persistence to four decimal places (RMSE 0.0006\,$^\circ$C), consistent with little temperature change within 0.1\,s. At $h=10$ and 30, the adapted model gives the lowest reported RMSE (0.0018 and 0.0032\,$^\circ$C), approximately 18\,\% and 32\,\% lower than the Transformer, the strongest measured-input baseline at these horizons. In contrast, the proposed model without adaptation is the weakest learned model, with RMSEs of 0.0026 and 0.0057\,$^\circ$C. Adaptation increases total test time substantially; at $h=10$, it increases from 321.30 to 4612.88\,s. These are descriptive comparisons at shorter physical forecast durations and a different training budget, not statistical evidence of lower prediction errors at the original sampling interval or a controlled comparison of sampling intervals at a fixed physical horizon.

\begin{table}[H]
\centering
\caption{Baseline comparison at the original sampling interval $\Delta t=0.1$\,s (single seed).}
\label{tab:dt01_drive_cycle}

\resizebox{\textwidth}{!}{%
\begin{tabular}{cl ccc c}
\toprule
\multirow{2}{*}{\textbf{\makecell{Horizon\\(steps)}}} &
\multirow{2}{*}{\textbf{Method}} &
\multicolumn{3}{c}{\textbf{Temperature Prediction ($^\circ$C)}} &
\multirow{2}{*}{\textbf{\makecell{Test time\\(s)}}} \\
\cmidrule(lr){3-5}
& &
\textbf{RMSE $\downarrow$} & \textbf{\(R^2\) $\uparrow$} & \textbf{MAE $\downarrow$} & \\
\midrule

\multirow{7}{*}{1 (0.1\,s)}
& Persistence & 0.0006 & -- & -- & -- \\
& XGBoost (I, V, T) & 0.0006 & 1.0000 & 0.0002 & 2.55 \\
& LSTM (I, V, T) & 0.0006 & 1.0000 & 0.0002 & 2.78 \\
& GRU (I, V, T) & 0.0006 & 1.0000 & 0.0002 & 2.36 \\
& Transformer (I, V, T) & 0.0006 & 1.0000 & 0.0002 & 5.55 \\
\cmidrule(lr){2-6}
& Proposed & 0.0006 & 1.0000 & 0.0003 & 323.48 \\
& Proposed + LoRA & 0.0006 & 1.0000 & 0.0003 & 4665.59 \\
\midrule

\multirow{7}{*}{10 (1\,s)}
& Persistence & 0.0025 & -- & -- & -- \\
& XGBoost (I, V, T) & 0.0023 & 1.0000 & 0.0015 & 2.58 \\
& LSTM (I, V, T) & 0.0023 & 1.0000 & 0.0015 & 2.78 \\
& GRU (I, V, T) & 0.0023 & 1.0000 & 0.0015 & 2.38 \\
& Transformer (I, V, T) & 0.0022 & 1.0000 & 0.0015 & 5.56 \\
\cmidrule(lr){2-6}
& Proposed & 0.0026 & 1.0000 & 0.0021 & 321.30 \\
& Proposed + LoRA & \textbf{0.0018} & 1.0000 & \textbf{0.0013} & 4612.88 \\
\midrule

\multirow{7}{*}{30 (3\,s)}
& Persistence & 0.0060 & -- & -- & -- \\
& XGBoost (I, V, T) & 0.0053 & 1.0000 & 0.0035 & 2.58 \\
& LSTM (I, V, T) & 0.0051 & 1.0000 & 0.0034 & 2.78 \\
& GRU (I, V, T) & 0.0051 & 1.0000 & 0.0034 & 2.40 \\
& Transformer (I, V, T) & 0.0047 & 1.0000 & 0.0031 & 5.50 \\
\cmidrule(lr){2-6}
& Proposed & 0.0057 & 1.0000 & 0.0039 & 337.53 \\
& Proposed + LoRA & \textbf{0.0032} & 1.0000 & \textbf{0.0023} & 4622.02 \\

\bottomrule
\end{tabular}%
}
\end{table}

\subsection{Discussion and limitations}
\label{sec:exp_discussion}
Overall, the selected configuration with LoRA gives the lowest reported temperature RMSE at all three main horizons on the first dataset and at the two shorter main horizons on the second dataset. The second-dataset advantage over all learned baselines is statistically established only at the shortest main horizon, while the gains from online adaptation relative to the same unadapted architecture are significant at all three main horizons on the second dataset. Without online adaptation, the proposed model is not significantly better than the Transformer at $h=10$ and 30 on the first dataset, and it is worse than the strongest baseline at $h=10$ and 30 on the second dataset. The advantage at the longer horizons therefore comes mainly from online adaptation, which is part of the proposed framework but is not applied to the learned baselines. Relative to persistence, the adapted model reduces RMSE by 46\%, 76\% and 69\% at $h=1$, 10 and 30 on the first dataset and by 64\%, 21\% and 6\% on the second. At $h=1$, the RMSEs of all learned methods lie within 0.017\,$^\circ$C of each other on the first dataset and within 0.0025\,$^\circ$C on the second, so the differences between methods are larger in absolute terms at the longer horizons. The transfer and validation-MSE results further indicate that cross-dataset weight reuse and retention of prior predictive performance remain limitations of the evaluated approach.

Several further limitations should be noted. First, the configuration choices on the first dataset are based on its test-cell results, so the first-dataset gains may be optimistic; the second dataset, on which no configuration is tuned, is the more independent evaluation. Second, the second-dataset partition is profile-disjoint but not fully cell-disjoint. Third, online adaptation increases test time considerably, for example from 5.34 to 85.58\,s on the first dataset at $h=1$, and from 321.30 to 4612.88\,s on the second dataset at $\Delta t=0.1$\,s and $h=10$. Fourth, the baselines are trained with a single seed, so seed variability is quantified only for the proposed model. Fifth, the learned baselines are not adapted online, so the comparison does not show whether the same adaptation would give similar gains for these architectures. Sixth, published hybrid ECM--neural methods, such as that of Yang et al.~\cite{yang2026physics}, are discussed but not re-implemented, so the comparison is limited to measured-input baselines. Seventh, the state estimator starts from the reference initial SOC of each sequence; the effect of an initial SOC error on the estimated states and the temperature forecasts is not evaluated. Finally, the study predicts the surface temperature of single cells; pack-level behaviour and long-term ageing are not considered.

%% file: appendix.tex
\section{Additional results}
\label{app:ablations}
This appendix reports the detailed baseline results and the detailed results of the ablation studies summarised in Sections~\ref{sec:first_dataset} and~\ref{sec:second_dataset}. The experimental settings follow Section~\ref{sec:experiment_implementation}.

\subsection{Detailed baseline comparison}
\label{app:baselines}
Tables~\ref{tab:baselines_lco} and~\ref{tab:baselines_drive_cycle} extend the main comparison in Table~\ref{tab:main_comparison} with bootstrap standard deviations, $R^2$ and test times.

\begin{table}[H]
\centering
\caption{Baseline comparison on the first dataset. $\star$: significantly better than all four learned baselines after Holm correction.}
\label{tab:baselines_lco}

\resizebox{\textwidth}{!}{%
\begin{tabular}{cl ccc c}
\toprule
\multirow{2}{*}{\textbf{\makecell{Horizon\\(steps)}}} &
\multirow{2}{*}{\textbf{Method}} &
\multicolumn{3}{c}{\textbf{Temperature Prediction ($^\circ$C)}} &
\multirow{2}{*}{\textbf{\makecell{Test time\\(s)}}} \\
\cmidrule(lr){3-5}
& &
\textbf{RMSE $\downarrow$} & \textbf{\(R^2\) $\uparrow$} & \textbf{MAE $\downarrow$} & \\
\midrule

\multirow{7}{*}{1 (10\,s)}
& Persistence &
0.0910 & -- & -- & -- \\
& XGBoost (I, V, T) &
0.0655 $\pm$ 0.0007 & 1.0000 $\pm$ 0.0000 & 0.0402 $\pm$ 0.0002 & 0.05 \\
& LSTM (I, V, T) &
0.0559 $\pm$ 0.0003 & 1.0000 $\pm$ 0.0000 & 0.0369 $\pm$ 0.0002 & 0.06 \\
& GRU (I, V, T) &
0.0516 $\pm$ 0.0003 & 1.0000 $\pm$ 0.0000 & 0.0344 $\pm$ 0.0002 & 0.05 \\
& Transformer (I, V, T) &
0.0513 $\pm$ 0.0003 & 1.0000 $\pm$ 0.0000 & 0.0340 $\pm$ 0.0002 & 0.10 \\
\cmidrule(lr){2-6}
& Proposed &
0.0499 $\pm$ 0.0003$^\star$ & 1.0000 $\pm$ 0.0000 & 0.0334 $\pm$ 0.0001 & 5.21 \\
& Proposed + LoRA &
\textbf{0.0491 $\pm$ 0.0003}$^\star$ & 1.0000 $\pm$ 0.0000 & \textbf{0.0329 $\pm$ 0.0001} & 84.53 \\
\midrule

\multirow{7}{*}{10 (100\,s)}
& Persistence &
0.7122 & -- & -- & -- \\
& XGBoost (I, V, T) &
0.3486 $\pm$ 0.0075 & 0.9988 $\pm$ 0.0001 & 0.1378 $\pm$ 0.0012 & 0.05 \\
& LSTM (I, V, T) &
0.2315 $\pm$ 0.0042 & 0.9995 $\pm$ 0.0000 & 0.1161 $\pm$ 0.0008 & 0.05 \\
& GRU (I, V, T) &
0.2071 $\pm$ 0.0045 & 0.9996 $\pm$ 0.0000 & 0.1076 $\pm$ 0.0007 & 0.05 \\
& Transformer (I, V, T) &
0.1910 $\pm$ 0.0042 & 0.9996 $\pm$ 0.0000 & 0.1009 $\pm$ 0.0006 & 0.10 \\
\cmidrule(lr){2-6}
& Proposed &
0.1951 $\pm$ 0.0044 & 0.9996 $\pm$ 0.0000 & 0.1001 $\pm$ 0.0007 & 5.28 \\
& Proposed + LoRA &
\textbf{0.1703 $\pm$ 0.0047}$^\star$ & 0.9997 $\pm$ 0.0000 & \textbf{0.0885 $\pm$ 0.0006} & 85.31 \\
\midrule

\multirow{7}{*}{30 (300\,s)}
& Persistence &
1.8637 & -- & -- & -- \\
& XGBoost (I, V, T) &
0.9405 $\pm$ 0.0157 & 0.9915 $\pm$ 0.0003 & 0.3301 $\pm$ 0.0036 & 0.05 \\
& LSTM (I, V, T) &
0.7614 $\pm$ 0.0124 & 0.9944 $\pm$ 0.0002 & 0.3018 $\pm$ 0.0028 & 0.05 \\
& GRU (I, V, T) &
0.7418 $\pm$ 0.0120 & 0.9947 $\pm$ 0.0002 & 0.2948 $\pm$ 0.0027 & 0.04 \\
& Transformer (I, V, T) &
0.6730 $\pm$ 0.0134 & 0.9956 $\pm$ 0.0002 & 0.2703 $\pm$ 0.0025 & 0.10 \\
\cmidrule(lr){2-6}
& Proposed &
0.6380 $\pm$ 0.0120 & 0.9961 $\pm$ 0.0001 & 0.2546 $\pm$ 0.0024 & 5.20 \\
& Proposed + LoRA &
\textbf{0.5867 $\pm$ 0.0125}$^\star$ & 0.9967 $\pm$ 0.0001 & \textbf{0.2297 $\pm$ 0.0022} & 81.59 \\

\bottomrule
\end{tabular}%
}
\end{table}

\begin{table}[H]
\centering
\caption{Baseline comparison on the second dataset. $\star$: significantly better than all four learned baselines after Holm correction.}
\label{tab:baselines_drive_cycle}

\resizebox{\textwidth}{!}{%
\begin{tabular}{cl ccc c}
\toprule
\multirow{2}{*}{\textbf{\makecell{Horizon\\(steps)}}} &
\multirow{2}{*}{\textbf{Method}} &
\multicolumn{3}{c}{\textbf{Temperature Prediction ($^\circ$C)}} &
\multirow{2}{*}{\textbf{\makecell{Test time\\(s)}}} \\
\cmidrule(lr){3-5}
& &
\textbf{RMSE $\downarrow$} & \textbf{\(R^2\) $\uparrow$} & \textbf{MAE $\downarrow$} & \\
\midrule

\multirow{7}{*}{1 (10\,s)}
& Persistence & 0.0183 & -- & -- & -- \\
& XGBoost (I, V, T) & 0.0074 $\pm$ 0.0001 & 1.0000 $\pm$ 0.0000 & 0.0052 $\pm$ 0.0000 & 0.03 \\
& LSTM (I, V, T) & 0.0090 $\pm$ 0.0001 & 1.0000 $\pm$ 0.0000 & 0.0065 $\pm$ 0.0000 & 0.03 \\
& GRU (I, V, T) & 0.0089 $\pm$ 0.0001 & 1.0000 $\pm$ 0.0000 & 0.0065 $\pm$ 0.0000 & 0.03 \\
& Transformer (I, V, T) & 0.0087 $\pm$ 0.0001 & 1.0000 $\pm$ 0.0000 & 0.0062 $\pm$ 0.0000 & 0.06 \\
\cmidrule(lr){2-6}
& Proposed & 0.0070 $\pm$ 0.0001$^\star$ & 1.0000 $\pm$ 0.0000 & 0.0049 $\pm$ 0.0000 & 3.30 \\
& Proposed + LoRA & \textbf{0.0065 $\pm$ 0.0001}$^\star$ & 1.0000 $\pm$ 0.0000 & \textbf{0.0045 $\pm$ 0.0000} & 46.71 \\
\midrule

\multirow{7}{*}{10 (100\,s)}
& Persistence & 0.1306 & -- & -- & -- \\
& XGBoost (I, V, T) & 0.1134 $\pm$ 0.0010 & 0.9999 $\pm$ 0.0000 & 0.0745 $\pm$ 0.0005 & 0.03 \\
& LSTM (I, V, T) & 0.1491 $\pm$ 0.0012 & 0.9999 $\pm$ 0.0000 & 0.0985 $\pm$ 0.0006 & 0.03 \\
& GRU (I, V, T) & 0.1479 $\pm$ 0.0011 & 0.9999 $\pm$ 0.0000 & 0.1001 $\pm$ 0.0006 & 0.03 \\
& Transformer (I, V, T) & 0.1100 $\pm$ 0.0009 & 0.9999 $\pm$ 0.0000 & 0.0731 $\pm$ 0.0004 & 0.07 \\
\cmidrule(lr){2-6}
& Proposed & 0.1163 $\pm$ 0.0010 & 0.9999 $\pm$ 0.0000 & 0.0785 $\pm$ 0.0005 & 3.31 \\
& Proposed + LoRA & \textbf{0.1037 $\pm$ 0.0011} & 0.9999 $\pm$ 0.0000 & \textbf{0.0679 $\pm$ 0.0004} & 47.02 \\
\midrule

\multirow{7}{*}{30 (300\,s)}
& Persistence & 0.2714 & -- & -- & -- \\
& XGBoost (I, V, T) & 0.2509 $\pm$ 0.0030 & 0.9997 $\pm$ 0.0000 & 0.1572 $\pm$ 0.0011 & 0.03 \\
& LSTM (I, V, T) & \textbf{0.2442 $\pm$ 0.0031} & 0.9997 $\pm$ 0.0000 & 0.1448 $\pm$ 0.0011 & 0.03 \\
& GRU (I, V, T) & 0.2452 $\pm$ 0.0032 & 0.9997 $\pm$ 0.0000 & \textbf{0.1427 $\pm$ 0.0011} & 0.03 \\
& Transformer (I, V, T) & 0.2583 $\pm$ 0.0034 & 0.9996 $\pm$ 0.0000 & 0.1519 $\pm$ 0.0011 & 0.07 \\
\cmidrule(lr){2-6}
& Proposed & 0.2650 $\pm$ 0.0036 & 0.9996 $\pm$ 0.0000 & 0.1596 $\pm$ 0.0012 & 3.29 \\
& Proposed + LoRA & 0.2556 $\pm$ 0.0038 & 0.9997 $\pm$ 0.0000 & 0.1484 $\pm$ 0.0011 & 45.90 \\

\bottomrule
\end{tabular}%
}
\end{table}

\subsection{Equivalent circuit model}
\label{app:ecm}
The ECM ablations use XGBoost with a lookback of 10 steps. Table~\ref{tab:ocv_soc_ablation} compares five OCV--SOC mappings for the 2-RC ECM. The exponential model gives the lowest temperature RMSE (0.0587\,$^\circ$C), whereas spline interpolation gives the lowest voltage RMSE (77.61\,mV). The SOC metrics are unchanged across the mappings. The exponential mapping is retained because temperature forecasting is the primary objective.

\begin{table}[H]
\centering
\caption{OCV--SOC mapping ablation on the first dataset (2-RC ECM and XGBoost).}
\label{tab:ocv_soc_ablation}

\resizebox{\textwidth}{!}{%
\begin{tabular}{lcccccccccc}
\toprule
\ecmheader{OCV--SOC Mapping}
\midrule

Polynomial fitting &
89.06 $\pm$ 17.27 & 0.95 $\pm$ 0.02 & 58.60 $\pm$ 16.98 &
5.31 $\pm$ 2.47 & 0.97 $\pm$ 0.02 & 4.44 $\pm$ 2.10 &
0.0597 $\pm$ 0.0004 & 1.0000 $\pm$ 0.0000 & 0.0391 $\pm$ 0.0002 &
\textbf{2.28} \\

Spline interpolation &
\textbf{77.61 $\pm$ 19.55} & \textbf{0.96 $\pm$ 0.02} & \textbf{47.67 $\pm$ 18.25} &
5.31 $\pm$ 2.47 & 0.97 $\pm$ 0.02 & 4.44 $\pm$ 2.10 &
0.0589 $\pm$ 0.0003 & 1.0000 $\pm$ 0.0000 & \textbf{0.0387 $\pm$ 0.0002} &
3.03 \\

Linear interpolation &
78.17 $\pm$ 19.24 & \textbf{0.96 $\pm$ 0.02} & 48.96 $\pm$ 18.98 &
5.31 $\pm$ 2.47 & 0.97 $\pm$ 0.02 & 4.44 $\pm$ 2.10 &
0.0596 $\pm$ 0.0003 & 1.0000 $\pm$ 0.0000 & 0.0392 $\pm$ 0.0002 &
2.49 \\

Exponential model &
77.98 $\pm$ 19.78 & \textbf{0.96 $\pm$ 0.02} & 47.89 $\pm$ 19.03 &
5.31 $\pm$ 2.47 & 0.97 $\pm$ 0.02 & 4.44 $\pm$ 2.10 &
\textbf{0.0587 $\pm$ 0.0003} & 1.0000 $\pm$ 0.0000 & \textbf{0.0387 $\pm$ 0.0002} &
2.36 \\

Logarithmic model &
84.80 $\pm$ 20.55 & \textbf{0.96 $\pm$ 0.02} & 53.75 $\pm$ 19.67 &
5.31 $\pm$ 2.47 & 0.97 $\pm$ 0.02 & 4.44 $\pm$ 2.10 &
0.0596 $\pm$ 0.0003 & 1.0000 $\pm$ 0.0000 & 0.0391 $\pm$ 0.0002 &
2.70 \\

\bottomrule
\end{tabular}%
}
\end{table}

With exponential OCV, Table~\ref{tab:rc_number_ablation} compares one, two and three RC networks. The respective temperature RMSEs are 0.0597, 0.0587 and 0.0588\,$^\circ$C. In the three-branch model, the second and third branches converge to identical parameters ($R_2=R_3=4.6\,\mathrm{m}\Omega$ and $C_2=C_3=6595$\,F), so the additional branch contributes no distinct dynamics. The 2-RC configuration is therefore retained.

\begin{table}[H]
\centering
\caption{Effect of the number of RC networks in the ECM.}
\label{tab:rc_number_ablation}

\resizebox{\textwidth}{!}{%
\begin{tabular}{lcccccccccc}
\toprule
\ecmheader{ECM Configuration}
\midrule

1-RC &
78.01 $\pm$ 19.76 & 0.96 $\pm$ 0.02 & 47.88 $\pm$ 19.04 &
5.31 $\pm$ 2.47 & 0.97 $\pm$ 0.02 & 4.44 $\pm$ 2.10 &
0.0597 $\pm$ 0.0004 & 1.0000 $\pm$ 0.0000 & 0.0390 $\pm$ 0.0002 &
\textbf{2.24} \\

2-RC &
\textbf{77.98 $\pm$ 19.78} & 0.96 $\pm$ 0.02 & 47.89 $\pm$ 19.03 &
5.31 $\pm$ 2.47 & 0.97 $\pm$ 0.02 & 4.44 $\pm$ 2.10 &
\textbf{0.0587 $\pm$ 0.0003} & 1.0000 $\pm$ 0.0000 & \textbf{0.0387 $\pm$ 0.0002} &
2.36 \\

3-RC &
78.23 $\pm$ 20.18 & 0.96 $\pm$ 0.02 & \textbf{47.79 $\pm$ 19.60} &
5.31 $\pm$ 2.47 & 0.97 $\pm$ 0.02 & 4.44 $\pm$ 2.10 &
0.0588 $\pm$ 0.0003 & 1.0000 $\pm$ 0.0000 & \textbf{0.0387 $\pm$ 0.0002} &
2.56 \\

\bottomrule
\end{tabular}%
}
\end{table}

Table~\ref{tab:hysteresis_ablation} evaluates the one-state hysteresis model of Plett~\cite{plett2004ekf2}, with fitted magnitude $M_H=0.066$\,V. Adding hysteresis decreases temperature RMSE from 0.0587 to 0.0578\,$^\circ$C and MAE from 0.0387 to 0.0377\,$^\circ$C, although voltage RMSE increases from 77.98 to 79.20\,mV. Hysteresis is retained for subsequent experiments.

\begin{table}[H]
\centering
\caption{Effect of hysteresis modelling in the 2-RC ECM.}
\label{tab:hysteresis_ablation}

\resizebox{\textwidth}{!}{%
\begin{tabular}{lcccccccccc}
\toprule
\ecmheader{Hysteresis Configuration}
\midrule

Without hysteresis &
\textbf{77.98 $\pm$ 19.78} & 0.96 $\pm$ 0.02 & 47.89 $\pm$ 19.03 &
5.31 $\pm$ 2.47 & 0.97 $\pm$ 0.02 & 4.44 $\pm$ 2.10 &
0.0587 $\pm$ 0.0003 & 1.0000 $\pm$ 0.0000 & 0.0387 $\pm$ 0.0002 &
\textbf{2.36} \\

With hysteresis &
79.20 $\pm$ 19.15 & 0.96 $\pm$ 0.02 & \textbf{45.14 $\pm$ 16.76} &
5.31 $\pm$ 2.47 & 0.97 $\pm$ 0.02 & 4.44 $\pm$ 2.10 &
\textbf{0.0578 $\pm$ 0.0004} & 1.0000 $\pm$ 0.0000 & \textbf{0.0377 $\pm$ 0.0002} &
2.76 \\

\bottomrule
\end{tabular}%
}
\end{table}

The electro-thermal variant makes $R_0$, $R_1$ and $R_2$ temperature-dependent through the Arrhenius relation $R(T)=R_{\mathrm{ref}}\exp[E(1/T-1/T_{\mathrm{ref}})]$, where temperatures are in kelvin, with reference temperature 25\,$^\circ$C and fitted coefficients $E_{R_0}=3605$\,K and $E_R=1759$\,K. Table~\ref{tab:thermal_coupling_ablation} shows that thermal coupling slightly improves voltage RMSE (79.20 to 78.93\,mV) but increases temperature RMSE (0.0578 to 0.0591\,$^\circ$C). Thus, the electrical-only ECM is used subsequently.

\begin{table}[H]
\centering
\caption{Electrical-only and electro-thermal ECM comparison.}
\label{tab:thermal_coupling_ablation}

\resizebox{\textwidth}{!}{%
\begin{tabular}{lcccccccccc}
\toprule
\ecmheader{Thermal Configuration}
\midrule

Electrical-only ECM &
79.20 $\pm$ 19.15 & 0.96 $\pm$ 0.02 & 45.14 $\pm$ 16.76 &
5.31 $\pm$ 2.47 & 0.97 $\pm$ 0.02 & 4.44 $\pm$ 2.10 &
\textbf{0.0578 $\pm$ 0.0004} & 1.0000 $\pm$ 0.0000 & \textbf{0.0377 $\pm$ 0.0002} &
\textbf{2.76} \\

Electro-thermal ECM &
\textbf{78.93 $\pm$ 15.00} & 0.96 $\pm$ 0.01 & \textbf{43.06 $\pm$ 13.87} &
5.31 $\pm$ 2.47 & 0.97 $\pm$ 0.02 & 4.44 $\pm$ 2.10 &
0.0591 $\pm$ 0.0004 & 1.0000 $\pm$ 0.0000 & 0.0387 $\pm$ 0.0002 &
2.89 \\

\bottomrule
\end{tabular}%
}
\end{table}

The state-estimation comparison (Table~\ref{tab:filter_ablation}) considers the Kalman filter (KF), extended Kalman filter (EKF), unscented Kalman filter (UKF) and particle filter (PF). The filter state contains SOC, RC voltages and hysteresis; voltage is evaluated before the measurement update. The KF uses a linear OCV fit, and the PF uses 500 particles. Relative to the unfiltered ECM, the EKF reduces voltage RMSE from 79.20 to 71.38\,mV, SOC RMSE from 5.31\,\% to 4.33\,\% and temperature RMSE from 0.0578 to 0.0574\,$^\circ$C. It gives the lowest RMSE for all three quantities among the evaluated filters, although it is not the fastest. The selected ECM therefore comprises two RC networks, exponential OCV, hysteresis and EKF-based state estimation, without explicit thermal coupling.

\begin{table}[H]
\centering
\caption{State-estimation filter comparison for the selected electrical ECM.}
\label{tab:filter_ablation}

\resizebox{\textwidth}{!}{%
\begin{tabular}{lcccccccccc}
\toprule
\ecmheader{Filtering Method}
\midrule

Kalman Filter (KF) &
121.63 $\pm$ 19.38 & 0.92 $\pm$ 0.03 & 57.86 $\pm$ 11.96 &
5.12 $\pm$ 1.19 & 0.98 $\pm$ 0.01 & 4.35 $\pm$ 0.95 &
0.0576 $\pm$ 0.0004 & 1.0000 $\pm$ 0.0000 & \textbf{0.0376 $\pm$ 0.0002} &
\textbf{5.42} \\

Extended Kalman Filter (EKF) &
\textbf{71.38 $\pm$ 16.82} & \textbf{0.97 $\pm$ 0.01} & \textbf{38.72 $\pm$ 11.02} &
\textbf{4.33 $\pm$ 1.57} & 0.98 $\pm$ 0.01 & \textbf{3.53 $\pm$ 1.27} &
\textbf{0.0574 $\pm$ 0.0003} & 1.0000 $\pm$ 0.0000 & 0.0378 $\pm$ 0.0002 &
8.44 \\

Unscented Kalman Filter (UKF) &
76.64 $\pm$ 19.37 & \textbf{0.97 $\pm$ 0.02} & 42.36 $\pm$ 12.57 &
4.41 $\pm$ 1.70 & 0.98 $\pm$ 0.01 & 3.56 $\pm$ 1.37 &
0.0575 $\pm$ 0.0003 & 1.0000 $\pm$ 0.0000 & 0.0378 $\pm$ 0.0002 &
9.51 \\

Particle Filter (PF) &
76.84 $\pm$ 19.00 & \textbf{0.97 $\pm$ 0.02} & 41.51 $\pm$ 11.38 &
4.44 $\pm$ 1.74 & 0.98 $\pm$ 0.01 & 3.60 $\pm$ 1.41 &
0.0576 $\pm$ 0.0004 & 1.0000 $\pm$ 0.0000 & 0.0377 $\pm$ 0.0002 &
10.63 \\

\bottomrule
\end{tabular}%
}
\end{table}

\subsection{Lookback window and prediction horizon}
\label{app:lookback}
Table~\ref{tab:lookback_ablation} reports decreasing temperature RMSE as the XGBoost lookback increases from 1 to 200 steps. The 200-step window (2000\,s) gives 0.0525\,$^\circ$C, compared with 0.0574\,$^\circ$C for the 10-step window in Table~\ref{tab:feature_ablation}. However, longer windows exclude the beginning of each cycle. On common test points ($t\geq 2000$\,s), the RMSEs for lookbacks of 1, 5, 50 and 200 steps are 0.0558, 0.0557, 0.0530 and 0.0525\,$^\circ$C, respectively. Thus, part of the apparent improvement with longer histories is attributable to the change in evaluated samples. A lookback of 10 steps is retained to avoid requiring long measurement histories.

\begin{table}[H]
\centering
\caption{Effect of lookback length with XGBoost and current-step ECM context.}
\label{tab:lookback_ablation}

\resizebox{\textwidth}{!}{%
\begin{tabular}{lcccc}
\toprule
\nnheader{Lookback (steps)}
\midrule

1 &
0.0593 $\pm$ 0.0003 & 1.0000 $\pm$ 0.0000 & 0.0393 $\pm$ 0.0002 &
\textbf{7.94} \\

5 &
0.0581 $\pm$ 0.0004 & 1.0000 $\pm$ 0.0000 & 0.0379 $\pm$ 0.0002 &
8.28 \\

50 &
0.0555 $\pm$ 0.0003 & 1.0000 $\pm$ 0.0000 & 0.0366 $\pm$ 0.0002 &
8.22 \\

200 &
\textbf{0.0525 $\pm$ 0.0004} & 1.0000 $\pm$ 0.0000 & \textbf{0.0330 $\pm$ 0.0002} &
8.46 \\

\bottomrule
\end{tabular}%
}
\end{table}

With this lookback, Table~\ref{tab:horizon_ablation} shows that XGBoost RMSE increases from 0.0575\,$^\circ$C at 10\,s to 0.2465\,$^\circ$C at 100\,s and 1.4043\,$^\circ$C at 600\,s. The one-step configuration corresponds to the selected input configuration in Table~\ref{tab:feature_ablation}; the small difference from 0.0574\,$^\circ$C is attributed to differences in the machine and library versions. Prediction time is 0.05\,s across the evaluated horizons, with the small variation in ECM timing attributed to run-to-run noise. The one-step horizon is used for the subsequent architecture ablations.

\begin{table}[H]
\centering
\caption{Effect of prediction horizon on XGBoost with ECM context.}
\label{tab:horizon_ablation}

\resizebox{\textwidth}{!}{%
\begin{tabular}{lcccccc}
\toprule
\nnheadersplit{Horizon (steps)}
\midrule

1 (10\,s) &
\textbf{0.0575 $\pm$ 0.0003} & \textbf{1.0000 $\pm$ 0.0000} & \textbf{0.0377 $\pm$ 0.0002} &
5.14 & 5.09 & 0.05 \\

3 (30\,s) &
0.0968 $\pm$ 0.0011 & 0.9999 $\pm$ 0.0000 & 0.0620 $\pm$ 0.0003 &
5.06 & 5.01 & 0.05 \\

5 (50\,s) &
0.1361 $\pm$ 0.0021 & 0.9998 $\pm$ 0.0000 & 0.0801 $\pm$ 0.0004 &
5.05 & 5.01 & 0.05 \\

10 (100\,s) &
0.2465 $\pm$ 0.0043 & 0.9994 $\pm$ 0.0000 & 0.1199 $\pm$ 0.0008 &
5.16 & 5.10 & 0.05 \\

15 (150\,s) &
0.3859 $\pm$ 0.0076 & 0.9986 $\pm$ 0.0001 & 0.1577 $\pm$ 0.0014 &
5.03 & 4.98 & 0.05 \\

30 (300\,s) &
0.8060 $\pm$ 0.0130 & 0.9937 $\pm$ 0.0002 & 0.2988 $\pm$ 0.0031 &
5.08 & 5.03 & 0.05 \\

60 (600\,s) &
1.4043 $\pm$ 0.0175 & 0.9810 $\pm$ 0.0005 & 0.5652 $\pm$ 0.0054 &
5.11 & 5.06 & 0.05 \\

\bottomrule
\end{tabular}%
}
\end{table}

\subsection{Recurrent hidden state}
\label{app:hidden}
The previous hidden embedding is incorporated into the iTransformer using either a linear transformation with ReLU or GRU gating. Let $z_t$ denote the iTransformer output and $c_t$ the current ECM context. The alternatives compute the 64-unit hidden state as
\begin{equation}
 h_t=\mathrm{ReLU}(W[z_t,c_t,h_{t-1}]+b)
\end{equation}
or
\begin{equation}
 h_t=\mathrm{GRUCell}([z_t,c_t],h_{t-1}).
\end{equation}
The state $h_t$ is used for temperature prediction and passed to the next step, with the cycle-wise reset and training protocol described in Section~\ref{sec:experiment_implementation}.

At lookback 10, GRU gating reduces RMSE from 0.0510 to 0.0498\,$^\circ$C, while the linear recurrent variant increases it to 0.0550\,$^\circ$C (Table~\ref{tab:hidden_state_ablation}). At lookback 1, GRU gating also improves RMSE, from 0.0571 to 0.0555\,$^\circ$C. The gated model with lookback 10 gives the lowest RMSE, although its prediction time increases from 0.05 to 0.22\,s relative to the non-recurrent iTransformer. It is retained for the continual-learning experiments. In a separate pair of runs of the gated model at lookback 10, removing the ECM context increased RMSE by 1.8\,\% (to 0.0510\,$^\circ$C) and MAE by 3.0\,\% (to 0.0345\,$^\circ$C) relative to the same model with the ECM context, while reducing the total test time from 5.20 to 0.22\,s; this difference was not tested for significance.

\begin{table}[H]
\centering
\caption{Effect of the recurrent hidden state at lookbacks of 10 and 1.}
\label{tab:hidden_state_ablation}

\resizebox{\textwidth}{!}{%
\begin{tabular}{clcccccc}
\toprule
\multirow{2}{*}{\textbf{Lookback}} &
\multirow{2}{*}{\textbf{Configuration}} &
\multicolumn{3}{c}{\textbf{Temperature Prediction ($^\circ$C)}} &
\multicolumn{3}{c}{\textbf{Time (s)}} \\
\cmidrule(lr){3-5}
\cmidrule(lr){6-8}
& &
\textbf{RMSE $\downarrow$} & \textbf{\(R^2\) $\uparrow$} & \textbf{MAE $\downarrow$} &
\textbf{Total} & \textbf{ECM} & \textbf{Prediction} \\
\midrule

\multirow{3}{*}{10}
& iTransformer, no hidden state &
0.0510 $\pm$ 0.0003 & 1.0000 $\pm$ 0.0000 & 0.0336 $\pm$ 0.0002 &
5.08 & 5.03 & 0.05 \\
& + $h_{t-1}$ (Linear + ReLU) &
0.0550 $\pm$ 0.0003 & 1.0000 $\pm$ 0.0000 & 0.0363 $\pm$ 0.0002 &
5.33 & 5.11 & 0.22 \\
& + $h_{t-1}$ (GRU gating) &
\textbf{0.0498 $\pm$ 0.0003} & 1.0000 $\pm$ 0.0000 & \textbf{0.0331 $\pm$ 0.0002} &
5.34 & 5.11 & 0.22 \\

\midrule

\multirow{3}{*}{1}
& iTransformer, no hidden state &
0.0571 $\pm$ 0.0003 & 1.0000 $\pm$ 0.0000 & 0.0382 $\pm$ 0.0002 &
5.10 & 5.07 & \textbf{0.03} \\
& + $h_{t-1}$ (Linear + ReLU) &
0.0568 $\pm$ 0.0003 & 1.0000 $\pm$ 0.0000 & 0.0380 $\pm$ 0.0002 &
5.26 & 5.05 & 0.21 \\
& + $h_{t-1}$ (GRU gating) &
0.0555 $\pm$ 0.0003 & 1.0000 $\pm$ 0.0000 & 0.0365 $\pm$ 0.0002 &
5.26 & 5.04 & 0.22 \\

\bottomrule
\end{tabular}%
}
\end{table}

\subsection{Prediction horizon with online adaptation}
\label{app:horizon_adapt}
Table~\ref{tab:horizon_adapt_ablation} evaluates the recurrent iTransformer with the relative-change target at horizons from 10 to 600\,s. A separate starting model is trained from scratch at each horizon; all such trainings run for at least 134 epochs on the machine equipped with the RTX 5060. The adaptation protocol follows Section~\ref{sec:continual_experiments}. The number of eligible test steps decreases from 63\,404 at $h=1$ to 58\,566 at $h=60$, since the last $h$ steps of a cycle lack an observed target. All methods within each horizon use the same test samples.

Persistence RMSE increases from 0.0910 to 2.8478\,$^\circ$C over the evaluated range. LoRA reduces the point-estimate RMSE relative to the unadapted model at every horizon. For example, RMSE decreases from 0.2003 to 0.1752\,$^\circ$C at 100\,s and from 0.7060 to 0.6398\,$^\circ$C at 300\,s. The reductions are significant for $h\leq30$. At $h=60$, the Wilcoxon test remains significant ($p=0.005$, with improvements in 57 of 82 cycles), but the bootstrap CI includes zero. LoRA also significantly outperforms XGBoost for $h\leq30$ ($p<10^{-4}$). At $h=60$, its RMSE is approximately 0.13\,$^\circ$C lower than XGBoost, but the difference is not significant ($p=0.065$, with improvements in 45 of 82 cycles).

\begin{figure}[H]
\centering
\includegraphics[width=\textwidth]{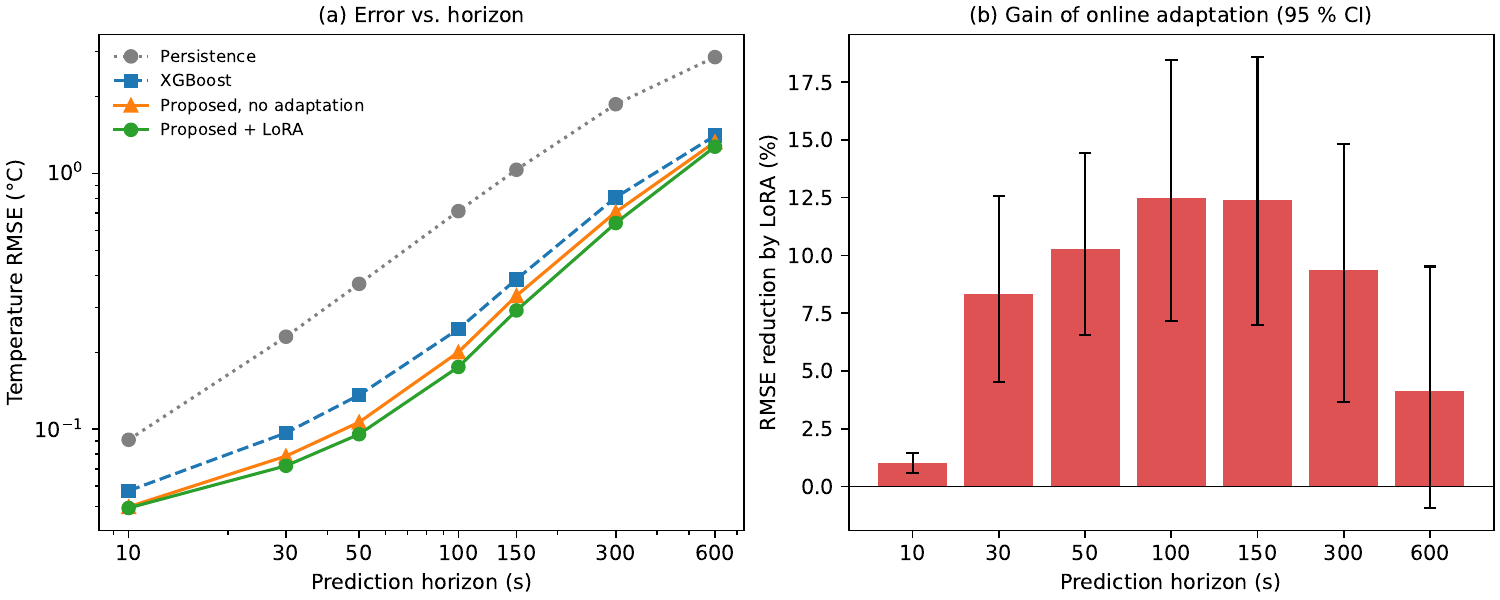}
\caption{Effect of horizon: (a) temperature RMSE (log--log); (b) relative RMSE reduction with LoRA (95\,\% bootstrap CI).}
\label{fig:horizon}
\end{figure}

The validation MSE increases after adaptation at several longer horizons, including an increase from 0.2539 to 0.5022 at $h=30$, indicating that improved test-cell forecasting can coincide with deterioration on the validation cell. Figure~\ref{fig:horizon} compares the horizon-dependent RMSEs and the relative reduction due to LoRA, with the largest relative gains at 100--150\,s. Prediction and online updates take approximately 1.24--1.27\,ms per step across the horizons in Table~\ref{tab:horizon_adapt_ablation}.

\begin{table}[H]
\centering
\caption{Horizon-dependent performance with the relative-change target. $\dagger$: significant improvement over no adaptation.}
\label{tab:horizon_adapt_ablation}

\resizebox{\textwidth}{!}{%
\begin{tabular}{cl cc ccc c cccc}
\toprule
\multirow{2}{*}{\textbf{\makecell{Horizon\\(steps)}}} &
\multirow{2}{*}{\textbf{Method}} &
\multicolumn{2}{c}{\textbf{Baselines RMSE ($^\circ$C)}} &
\multicolumn{3}{c}{\textbf{Temperature Prediction ($^\circ$C)}} &
\multirow{2}{*}{\textbf{\makecell{LCO\_4 MSE\\before $\rightarrow$ after}}} &
\multicolumn{4}{c}{\textbf{Time}} \\
\cmidrule(lr){3-4}
\cmidrule(lr){5-7}
\cmidrule(lr){9-12}
& &
\textbf{Persistence} & \textbf{XGBoost} &
\textbf{RMSE $\downarrow$} & \textbf{\(R^2\) $\uparrow$} & \textbf{MAE $\downarrow$} & &
\textbf{Total (s)} & \textbf{ECM (s)} & \textbf{Prediction (s)} & \textbf{\makecell{Prediction\\per step (ms)}} \\
\midrule

\multirow{2}{*}{1 (10\,s)}
& No adaptation & \multirow{2}{*}{0.0910} & \multirow{2}{*}{0.0575 $\pm$ 0.0003} &
0.0498 $\pm$ 0.0003 & 1.0000 $\pm$ 0.0000 & 0.0331 $\pm$ 0.0002 & 0.3748 &
5.34 & 5.11 & 0.22 & 0.003 \\
& + LoRA & & &
\textbf{0.0493 $\pm$ 0.0003}$^\dagger$ & 1.0000 $\pm$ 0.0000 & \textbf{0.0329 $\pm$ 0.0001} & 0.3748 $\rightarrow$ 0.3699 &
85.58 & 4.94 & 80.64 & 1.27 \\
\midrule

\multirow{2}{*}{3 (30\,s)}
& No adaptation & \multirow{2}{*}{0.2301} & \multirow{2}{*}{0.0968 $\pm$ 0.0011} &
0.0786 $\pm$ 0.0005 & 0.9999 $\pm$ 0.0000 & 0.0536 $\pm$ 0.0002 & 0.1278 &
5.18 & 4.96 & 0.22 & 0.003 \\
& + LoRA & & &
\textbf{0.0720 $\pm$ 0.0004}$^\dagger$ & 0.9999 $\pm$ 0.0000 & \textbf{0.0509 $\pm$ 0.0002} & 0.1278 $\rightarrow$ 0.1276 &
83.50 & 4.90 & 78.60 & 1.24 \\
\midrule

\multirow{2}{*}{5 (50\,s)}
& No adaptation & \multirow{2}{*}{0.3703} & \multirow{2}{*}{0.1361 $\pm$ 0.0021} &
0.1066 $\pm$ 0.0011 & 0.9999 $\pm$ 0.0000 & 0.0711 $\pm$ 0.0003 & 0.0908 &
5.18 & 4.96 & 0.22 & 0.003 \\
& + LoRA & & &
\textbf{0.0957 $\pm$ 0.0011}$^\dagger$ & 0.9999 $\pm$ 0.0000 & \textbf{0.0649 $\pm$ 0.0003} & 0.0908 $\rightarrow$ 0.1024 &
84.68 & 4.96 & 79.72 & 1.26 \\
\midrule

\multirow{2}{*}{10 (100\,s)}
& No adaptation & \multirow{2}{*}{0.7122} & \multirow{2}{*}{0.2465 $\pm$ 0.0043} &
0.2003 $\pm$ 0.0042 & 0.9996 $\pm$ 0.0000 & 0.1052 $\pm$ 0.0007 & 0.1110 &
5.15 & 4.93 & 0.22 & 0.004 \\
& + LoRA & & &
\textbf{0.1752 $\pm$ 0.0044}$^\dagger$ & 0.9997 $\pm$ 0.0000 & \textbf{0.0935 $\pm$ 0.0006} & 0.1110 $\rightarrow$ 0.1401 &
84.15 & 4.98 & 79.17 & 1.26 \\
\midrule

\multirow{2}{*}{15 (150\,s)}
& No adaptation & \multirow{2}{*}{1.0335} & \multirow{2}{*}{0.3859 $\pm$ 0.0076} &
0.3327 $\pm$ 0.0067 & 0.9989 $\pm$ 0.0000 & 0.1631 $\pm$ 0.0012 & 0.1511 &
5.13 & 4.91 & 0.22 & 0.004 \\
& + LoRA & & &
\textbf{0.2914 $\pm$ 0.0068}$^\dagger$ & 0.9992 $\pm$ 0.0000 & \textbf{0.1320 $\pm$ 0.0010} & 0.1511 $\rightarrow$ 0.1859 &
83.37 & 4.92 & 78.45 & 1.26 \\
\midrule

\multirow{2}{*}{30 (300\,s)}
& No adaptation & \multirow{2}{*}{1.8637} & \multirow{2}{*}{0.8060 $\pm$ 0.0130} &
0.7060 $\pm$ 0.0108 & 0.9952 $\pm$ 0.0001 & 0.2894 $\pm$ 0.0026 & 0.2539 &
5.15 & 4.94 & 0.22 & 0.004 \\
& + LoRA & & &
\textbf{0.6398 $\pm$ 0.0104}$^\dagger$ & 0.9960 $\pm$ 0.0001 & \textbf{0.2737 $\pm$ 0.0024} & 0.2539 $\rightarrow$ 0.5022 &
82.27 & 4.91 & 77.35 & 1.27 \\
\midrule

\multirow{2}{*}{60 (600\,s)}
& No adaptation & \multirow{2}{*}{2.8478} & \multirow{2}{*}{1.4043 $\pm$ 0.0175} &
1.3265 $\pm$ 0.0147 & 0.9830 $\pm$ 0.0004 & 0.5666 $\pm$ 0.0050 & 0.2909 &
5.19 & 4.98 & 0.21 & 0.004 \\
& + LoRA & & &
\textbf{1.2719 $\pm$ 0.0158} & 0.9844 $\pm$ 0.0004 & \textbf{0.5275 $\pm$ 0.0048} & 0.2909 $\rightarrow$ 0.3972 &
79.11 & 5.06 & 74.05 & 1.26 \\

\bottomrule
\end{tabular}%
}
\end{table}

\subsection{Physics-informed inputs and prediction target}
\label{app:target}
Table~\ref{tab:physics_ablation} examines whether explicitly derived thermal inputs or a thermal residual model improve the recurrent iTransformer. The heat-generation feature is computed from the ECM states and fitted resistances as
\begin{equation}
 Q_t=I_t^2R_0+\frac{U_{1,t}^2}{R_1}+\frac{U_{2,t}^2}{R_2}.
\end{equation}
The ambient-temperature variant appends $T_t-T_{\mathrm{amb}}$ to the ECM context, where $T_{\mathrm{amb}}$ is the climate-chamber temperature. The hybrid residual model uses the lumped thermal equation
\begin{equation}
 C\frac{\mathrm{d}T}{\mathrm{d}t}=Q-hA(T-T_{\mathrm{amb}}),
\end{equation}
with parameters fitted on the training cells only ($C=97.5$\,J/K, $hA=0.103$\,W/K and $\tau=944$\,s). Since future current is unknown, $Q$ is held constant over the forecast horizon, and the neural network predicts the residual relative to the lumped-model forecast. The standalone lumped model achieves RMSEs of 0.0695 and 0.4568\,$^\circ$C at $h=1$ and 10, respectively. Here $hA$ denotes the heat-transfer coefficient-area product, as distinct from the prediction horizon $h$.

At $h=1$, adding heat generation, the ambient-temperature difference or the lumped residual model increases RMSE from 0.0498 to 0.0501, 0.0507 or 0.0516\,$^\circ$C, respectively. At $h=10$, adding heat generation reduces the point estimate from 0.2003 to 0.1942\,$^\circ$C, but not significantly ($p=0.53$, 43 of 82 cycles improved); the lumped residual model gives 0.2316\,$^\circ$C. All trainings in this ablation run for at least 100 epochs; the lumped residual model at $h=10$ stops at this minimum.

Changing the relative-change target to the absolute temperature change $\Delta T=T_{t+h}-T_t$ gives RMSEs of 0.0499, 0.1951 and 0.6380\,$^\circ$C without adaptation, and 0.0491, 0.1703 and 0.5867\,$^\circ$C with LoRA, at $h=1$, 10 and 30, respectively. Relative to the corresponding relative-target configurations, the unadjusted joint significance criteria are met for $\Delta T$ with LoRA at $h=1$ and 10, and without adaptation at $h=10$ and 30. At $h=30$, the reduction with LoRA is 8.3\,\%, but is not significant ($p=0.097$, 46 of 82 cycles improved). Combining heat generation with $\Delta T$ at $h=10$ gives RMSEs of 0.1927 without adaptation and 0.1718\,$^\circ$C with LoRA; neither significantly improves upon the corresponding relative-target reference. The best adapted point estimates at the three horizons are obtained using $\Delta T$ without the additional thermal inputs. The term ``absolute target'' in Table~\ref{tab:physics_ablation} refers to the absolute change, not to direct prediction of temperature.

\begin{table}[H]
\centering
\caption{Physics-input and target ablation. $\dagger$: significant improvement over the relative-target reference with matching horizon and adaptation.}
\label{tab:physics_ablation}

\resizebox{\textwidth}{!}{%
\begin{tabular}{clcccccccc}
\toprule
\multirow{2}{*}{\textbf{\makecell{Horizon\\(steps)}}} &
\multirow{2}{*}{\textbf{Method}} &
\multicolumn{3}{c}{\textbf{Temperature Prediction ($^\circ$C)}} &
\multirow{2}{*}{\textbf{\makecell{LCO\_4 MSE\\before $\rightarrow$ after}}} &
\multicolumn{4}{c}{\textbf{Time}} \\
\cmidrule(lr){3-5}
\cmidrule(lr){7-10}
& &
\textbf{RMSE $\downarrow$} & \textbf{\(R^2\) $\uparrow$} & \textbf{MAE $\downarrow$} & &
\textbf{Total (s)} & \textbf{ECM (s)} & \textbf{Prediction (s)} & \textbf{\makecell{Prediction\\per step (ms)}} \\
\midrule

\multirow{7}{*}{1 (10\,s)}
& Reference (relative target), no adaptation &
\textbf{0.0498 $\pm$ 0.0003} & 1.0000 $\pm$ 0.0000 & \textbf{0.0331 $\pm$ 0.0002} & 0.3748 &
5.34 & 5.11 & 0.22 & 0.003 \\
& + heat generation $Q$ &
0.0501 $\pm$ 0.0003 & 1.0000 $\pm$ 0.0000 & 0.0334 $\pm$ 0.0002 & 0.3762 &
5.30 & 5.07 & 0.23 & 0.004 \\
& + $T - T_\mathrm{amb}$ &
0.0507 $\pm$ 0.0003 & 1.0000 $\pm$ 0.0000 & 0.0341 $\pm$ 0.0002 & 0.3757 &
5.31 & 5.09 & 0.22 & 0.003 \\
& Lumped thermal model + residual &
0.0516 $\pm$ 0.0003 & 1.0000 $\pm$ 0.0000 & 0.0349 $\pm$ 0.0002 & 0.5487 &
5.25 & 5.02 & 0.23 & 0.004 \\
& Absolute target &
0.0499 $\pm$ 0.0003 & 1.0000 $\pm$ 0.0000 & 0.0334 $\pm$ 0.0001 & 0.3078 &
5.21 & 4.98 & 0.23 & 0.004 \\
\cmidrule(lr){2-10}
& Reference + LoRA &
0.0493 $\pm$ 0.0003 & 1.0000 $\pm$ 0.0000 & 0.0329 $\pm$ 0.0001 & 0.3748 $\rightarrow$ 0.3699 &
85.58 & 4.94 & 80.64 & 1.27 \\
& Absolute target + LoRA &
\textbf{0.0491 $\pm$ 0.0003}$^\dagger$ & 1.0000 $\pm$ 0.0000 & \textbf{0.0329 $\pm$ 0.0001} & 0.3078 $\rightarrow$ 0.3138 &
84.53 & 5.27 & 79.27 & 1.25 \\
\midrule

\multirow{8}{*}{10 (100\,s)}
& Reference (relative target), no adaptation &
0.2003 $\pm$ 0.0042 & 0.9996 $\pm$ 0.0000 & 0.1052 $\pm$ 0.0007 & 0.1110 &
5.15 & 4.93 & 0.22 & 0.004 \\
& + heat generation $Q$ &
0.1942 $\pm$ 0.0041 & 0.9996 $\pm$ 0.0000 & 0.1042 $\pm$ 0.0006 & 0.0996 &
5.24 & 5.02 & 0.23 & 0.004 \\
& Lumped thermal model + residual &
0.2316 $\pm$ 0.0041 & 0.9995 $\pm$ 0.0000 & 0.1360 $\pm$ 0.0007 & 0.2312 &
5.36 & 5.14 & 0.23 & 0.004 \\
& Absolute target &
0.1951 $\pm$ 0.0044$^\dagger$ & 0.9996 $\pm$ 0.0000 & \textbf{0.1001 $\pm$ 0.0007} & 0.0589 &
5.28 & 5.05 & 0.23 & 0.004 \\
& $Q$ + absolute target &
\textbf{0.1927 $\pm$ 0.0043} & 0.9996 $\pm$ 0.0000 & 0.1026 $\pm$ 0.0006 & 0.0678 &
5.31 & 5.09 & 0.22 & 0.004 \\
\cmidrule(lr){2-10}
& Reference + LoRA &
0.1752 $\pm$ 0.0044 & 0.9997 $\pm$ 0.0000 & 0.0935 $\pm$ 0.0006 & 0.1110 $\rightarrow$ 0.1401 &
84.15 & 4.98 & 79.17 & 1.26 \\
& Absolute target + LoRA &
\textbf{0.1703 $\pm$ 0.0047}$^\dagger$ & 0.9997 $\pm$ 0.0000 & \textbf{0.0885 $\pm$ 0.0006} & 0.0589 $\rightarrow$ 0.0700 &
85.31 & 5.00 & 80.31 & 1.28 \\
& $Q$ + absolute target + LoRA &
0.1718 $\pm$ 0.0045 & 0.9997 $\pm$ 0.0000 & 0.0909 $\pm$ 0.0006 & 0.0678 $\rightarrow$ 0.0811 &
83.67 & 5.07 & 78.61 & 1.25 \\
\midrule

\multirow{5}{*}{30 (300\,s)}
& Reference (relative target), no adaptation &
0.7060 $\pm$ 0.0108 & 0.9952 $\pm$ 0.0001 & 0.2894 $\pm$ 0.0026 & 0.2539 &
5.15 & 4.94 & 0.22 & 0.004 \\
& Absolute target &
0.6380 $\pm$ 0.0120$^\dagger$ & 0.9961 $\pm$ 0.0001 & 0.2546 $\pm$ 0.0024 & 0.1002 &
5.20 & 4.98 & 0.22 & 0.004 \\
\cmidrule(lr){2-10}
& Reference + LoRA &
0.6398 $\pm$ 0.0104 & 0.9960 $\pm$ 0.0001 & 0.2737 $\pm$ 0.0024 & 0.2539 $\rightarrow$ 0.5022 &
82.27 & 4.91 & 77.35 & 1.27 \\
& Absolute target + LoRA &
\textbf{0.5867 $\pm$ 0.0125} & 0.9967 $\pm$ 0.0001 & \textbf{0.2297 $\pm$ 0.0022} & 0.1002 $\rightarrow$ 0.1551 &
81.59 & 5.06 & 76.53 & 1.25 \\

\bottomrule
\end{tabular}%
}
\end{table}

Table~\ref{tab:target_comparison} extends the comparison to direct prediction of $T_{t+h}$, alongside relative change and $\Delta T$. The direct target is significantly worse than $\Delta T$ at all three horizons, both with and without adaptation, following Holm correction over 12 comparisons. At $h=1$, direct prediction gives 0.2784\,$^\circ$C without adaptation and 0.3920\,$^\circ$C with LoRA, compared with 0.0499 and 0.0491\,$^\circ$C for $\Delta T$. The relative target is significantly worse than $\Delta T$ without adaptation at $h=10$ and 30, by 2.6\,\% and 10.7\,\%, respectively. After correction, the differences between the two change targets are not significant at $h=1$ or with LoRA at any horizon. This distinction is important because the earlier unadjusted comparisons in Table~\ref{tab:physics_ablation} identify some differences that do not remain significant after multiplicity correction. Based on the target comparison and the adapted point estimates, the final configuration uses the electrical 2-RC ECM with exponential OCV, hysteresis and EKF, the iTransformer with GRU hidden state, a 10-step measured-input window and the $\Delta T$ target, with LoRA as the online-adaptation variant.

\begin{table}[H]
\centering
\caption{Comparison of relative-change, absolute-change and direct-temperature targets on the first dataset.}
\label{tab:target_comparison}

\resizebox{\textwidth}{!}{%
\begin{tabular}{cll ccc c}
\toprule
\multirow{2}{*}{\textbf{\makecell{Horizon\\(steps)}}} &
\multirow{2}{*}{\textbf{Target}} &
\multirow{2}{*}{\textbf{Adaptation}} &
\multicolumn{3}{c}{\textbf{Temperature Prediction ($^\circ$C)}} &
\multirow{2}{*}{\textbf{\makecell{Test time\\(s)}}} \\
\cmidrule(lr){4-6}
& & &
\textbf{RMSE $\downarrow$} & \textbf{\(R^2\) $\uparrow$} & \textbf{MAE $\downarrow$} & \\
\midrule

\multirow{6}{*}{1 (10\,s)}
& Relative & -- & 0.0498 $\pm$ 0.0003 & 1.0000 $\pm$ 0.0000 & 0.0331 $\pm$ 0.0002 & 5.34 \\
& $\Delta T$ & -- & 0.0499 $\pm$ 0.0003 & 1.0000 $\pm$ 0.0000 & 0.0334 $\pm$ 0.0001 & 5.21 \\
& Direct & -- & 0.2784 $\pm$ 0.0016 & 0.9992 $\pm$ 0.0000 & 0.1927 $\pm$ 0.0008 & 5.33 \\
\cmidrule(lr){2-7}
& Relative & LoRA & 0.0493 $\pm$ 0.0003 & 1.0000 $\pm$ 0.0000 & 0.0329 $\pm$ 0.0001 & 85.58 \\
& $\Delta T$ & LoRA & \textbf{0.0491 $\pm$ 0.0003} & 1.0000 $\pm$ 0.0000 & \textbf{0.0329 $\pm$ 0.0001} & 84.53 \\
& Direct & LoRA & 0.3920 $\pm$ 0.0023 & 0.9985 $\pm$ 0.0000 & 0.2430 $\pm$ 0.0012 & 85.78 \\
\midrule

\multirow{6}{*}{10 (100\,s)}
& Relative & -- & 0.2003 $\pm$ 0.0042 & 0.9996 $\pm$ 0.0000 & 0.1052 $\pm$ 0.0007 & 5.15 \\
& $\Delta T$ & -- & 0.1951 $\pm$ 0.0044 & 0.9996 $\pm$ 0.0000 & 0.1001 $\pm$ 0.0007 & 5.28 \\
& Direct & -- & 0.5581 $\pm$ 0.0033 & 0.9970 $\pm$ 0.0000 & 0.4044 $\pm$ 0.0015 & 5.29 \\
\cmidrule(lr){2-7}
& Relative & LoRA & 0.1752 $\pm$ 0.0044 & 0.9997 $\pm$ 0.0000 & 0.0935 $\pm$ 0.0006 & 84.15 \\
& $\Delta T$ & LoRA & \textbf{0.1703 $\pm$ 0.0047} & 0.9997 $\pm$ 0.0000 & \textbf{0.0885 $\pm$ 0.0006} & 85.31 \\
& Direct & LoRA & 0.4150 $\pm$ 0.0034 & 0.9983 $\pm$ 0.0000 & 0.2598 $\pm$ 0.0013 & 84.02 \\
\midrule

\multirow{6}{*}{30 (300\,s)}
& Relative & -- & 0.7060 $\pm$ 0.0108 & 0.9952 $\pm$ 0.0001 & 0.2894 $\pm$ 0.0026 & 5.15 \\
& $\Delta T$ & -- & 0.6380 $\pm$ 0.0120 & 0.9961 $\pm$ 0.0001 & 0.2546 $\pm$ 0.0024 & 5.20 \\
& Direct & -- & 0.8649 $\pm$ 0.0130 & 0.9928 $\pm$ 0.0002 & 0.4580 $\pm$ 0.0030 & 5.20 \\
\cmidrule(lr){2-7}
& Relative & LoRA & 0.6398 $\pm$ 0.0104 & 0.9960 $\pm$ 0.0001 & 0.2737 $\pm$ 0.0024 & 82.27 \\
& $\Delta T$ & LoRA & \textbf{0.5867 $\pm$ 0.0125} & 0.9967 $\pm$ 0.0001 & \textbf{0.2297 $\pm$ 0.0022} & 81.59 \\
& Direct & LoRA & 0.7756 $\pm$ 0.0121 & 0.9942 $\pm$ 0.0002 & 0.3797 $\pm$ 0.0026 & 82.99 \\

\bottomrule
\end{tabular}%
}
\end{table}

\subsection{Application to the second dataset}
\label{app:second}
The selected ECM settings are applied to the second dataset, with parameters fitted on its training profiles (fit time 284\,s). Table~\ref{tab:drive_cycle_ecm} compares voltage and SOC estimates and one-step XGBoost temperature forecasts across datasets. Voltage RMSE is lower on the second dataset (57.93 versus 71.38\,mV), as are SOC RMSE (1.12 versus 4.33\,\%) and temperature RMSE (0.0076 versus 0.0575\,$^\circ$C). Persistence RMSE is also lower on the second dataset (0.0183 versus 0.0910\,$^\circ$C). The second-dataset profiles exhibit relatively slow temperature changes (1--7\,$^\circ$C per profile) and small sensor scatter (0.003\,$^\circ$C), consistent with the lower absolute temperature-prediction errors. These errors should therefore not be interpreted as directly comparable measures of forecasting difficulty across datasets.

\begin{table}[H]
\centering
\caption{Cross-dataset comparison of ECM estimation and one-step XGBoost temperature forecasting.}
\label{tab:drive_cycle_ecm}

\resizebox{\textwidth}{!}{%
\begin{tabular}{lcccccccccc}
\toprule
\ecmheader{Dataset}
\midrule

First (LCO, 82 test cycles) &
71.38 $\pm$ 16.82 & 0.97 $\pm$ 0.01 & 38.72 $\pm$ 11.02 &
4.33 $\pm$ 1.57 & 0.98 $\pm$ 0.01 & 3.53 $\pm$ 1.27 &
0.0575 $\pm$ 0.0003 & 1.0000 $\pm$ 0.0000 & 0.0377 $\pm$ 0.0002 &
5.14 \\

Second (NMC, 25 test profiles) &
57.93 $\pm$ 9.77 & 0.95 $\pm$ 0.02 & 40.60 $\pm$ 6.71 &
1.12 $\pm$ 0.65 & 1.00 $\pm$ 0.00 & 1.02 $\pm$ 0.61 &
0.0076 $\pm$ 0.0001 & 1.0000 $\pm$ 0.0000 & 0.0054 $\pm$ 0.0000 &
2.80 \\

\bottomrule
\end{tabular}%
}
\end{table}

Table~\ref{tab:drive_cycle_horizon} evaluates the selected architecture trained on the second dataset, using the same ECM across configurations. The proposed models use $\Delta T$ and online adaptation is performed independently for NMC063 (ID 8) and NMC069 (IDs 9--12), processing profiles in the order of ambient temperatures 5, 15, 25, 35 and 45\,$^\circ$C. Samples are eligible for updates only after their target temperatures have been measured. The common test sets contain 34\,653, 34\,428 and 33\,928 steps at $h=1$, 10 and 30, respectively. Training stops after 125 epochs at $h=1$ and at the minimum of 100 epochs at $h=10$ and 30, for which the learning rate is reduced to $3.9\times10^{-6}$.

LoRA significantly improves on the unadapted architecture at each horizon, reducing RMSE from 0.0070 to 0.0065, from 0.1163 to 0.1037 and from 0.2650 to 0.2556\,$^\circ$C. Relative to ECM-enhanced XGBoost, the adapted model is significantly better at $h=1$ and 10, but not at $h=30$, where the CI includes zero despite $p<10^{-4}$ and improvements in 23 of 25 profiles. The unadapted model significantly outperforms ECM-enhanced XGBoost only at $h=1$. In Table~\ref{tab:drive_cycle_horizon}, $\dagger$ denotes a significant improvement over the unadapted proposed model and $\ddagger$ denotes a significant improvement over ECM-enhanced XGBoost. Adaptation requires 1.27--1.28\,ms of prediction and update time per step; total test time increases from 3.29--3.31\,s to 45.90--47.02\,s. Validation MSE increases after adaptation at all three horizons, most notably from 0.6000 to 1.2618 at $h=10$. This indicates a trade-off between adaptation to the test cells and retention of validation performance. Since the test set includes profiles from a training cell, these aggregate results cannot be attributed exclusively to previously unseen cells.

\begin{table}[H]
\centering
\caption{Horizon-dependent results on the second dataset. $\dagger$: significant improvement over no adaptation; $\ddagger$: significant improvement over ECM-enhanced XGBoost.}
\label{tab:drive_cycle_horizon}

\resizebox{\textwidth}{!}{%
\begin{tabular}{clc ccc c cccc}
\toprule
\multirow{2}{*}{\textbf{\makecell{Horizon\\(steps)}}} &
\multirow{2}{*}{\textbf{Method}} &
\multirow{2}{*}{\textbf{\makecell{Persistence\\RMSE ($^\circ$C)}}} &
\multicolumn{3}{c}{\textbf{Temperature Prediction ($^\circ$C)}} &
\multirow{2}{*}{\textbf{\makecell{Validation MSE\\before $\rightarrow$ after}}} &
\multicolumn{4}{c}{\textbf{Time}} \\
\cmidrule(lr){4-6}
\cmidrule(lr){8-11}
& & &
\textbf{RMSE $\downarrow$} & \textbf{\(R^2\) $\uparrow$} & \textbf{MAE $\downarrow$} & &
\textbf{Total (s)} & \textbf{ECM (s)} & \textbf{Prediction (s)} & \textbf{\makecell{Prediction\\per step (ms)}} \\
\midrule

\multirow{3}{*}{1 (10\,s)}
& XGBoost & \multirow{3}{*}{0.0183} &
0.0076 $\pm$ 0.0001 & 1.0000 $\pm$ 0.0000 & 0.0054 $\pm$ 0.0000 & -- &
2.80 & 2.77 & 0.03 & 0.001 \\
& Proposed & &
0.0070 $\pm$ 0.0001$^\ddagger$ & 1.0000 $\pm$ 0.0000 & 0.0049 $\pm$ 0.0000 & 0.6105 &
3.30 & 2.78 & 0.52 & 0.015 \\
& Proposed + LoRA & &
\textbf{0.0065 $\pm$ 0.0001}$^{\dagger\ddagger}$ & 1.0000 $\pm$ 0.0000 & \textbf{0.0045 $\pm$ 0.0000} & 0.6105 $\rightarrow$ 0.6420 &
46.71 & 2.75 & 43.96 & 1.27 \\
\midrule

\multirow{3}{*}{10 (100\,s)}
& XGBoost & \multirow{3}{*}{0.1306} &
0.1185 $\pm$ 0.0011 & 0.9999 $\pm$ 0.0000 & 0.0789 $\pm$ 0.0005 & -- &
2.84 & 2.82 & 0.03 & 0.001 \\
& Proposed & &
0.1163 $\pm$ 0.0010 & 0.9999 $\pm$ 0.0000 & 0.0785 $\pm$ 0.0005 & 0.6000 &
3.31 & 2.79 & 0.52 & 0.015 \\
& Proposed + LoRA & &
\textbf{0.1037 $\pm$ 0.0011}$^{\dagger\ddagger}$ & 0.9999 $\pm$ 0.0000 & \textbf{0.0679 $\pm$ 0.0004} & 0.6000 $\rightarrow$ 1.2618 &
47.02 & 2.84 & 44.17 & 1.28 \\
\midrule

\multirow{3}{*}{30 (300\,s)}
& XGBoost & \multirow{3}{*}{0.2714} &
0.2753 $\pm$ 0.0029 & 0.9996 $\pm$ 0.0000 & 0.1817 $\pm$ 0.0011 & -- &
2.82 & 2.79 & 0.03 & 0.001 \\
& Proposed & &
0.2650 $\pm$ 0.0036 & 0.9996 $\pm$ 0.0000 & 0.1596 $\pm$ 0.0012 & 0.7251 &
3.29 & 2.77 & 0.52 & 0.015 \\
& Proposed + LoRA & &
\textbf{0.2556 $\pm$ 0.0038}$^\dagger$ & 0.9997 $\pm$ 0.0000 & \textbf{0.1484 $\pm$ 0.0011} & 0.7251 $\rightarrow$ 0.8248 &
45.90 & 2.77 & 43.13 & 1.27 \\

\bottomrule
\end{tabular}%
}
\end{table}